\documentclass{article}
\usepackage[preprint]{neurips_2026}

\usepackage{amsmath,amsfonts,bm}

\def\eqref#1{equation~\ref{#1}}

\def\1{\bm{1}}

\DeclareMathAlphabet{\mathsfit}{\encodingdefault}{\sfdefault}{m}{sl}
\SetMathAlphabet{\mathsfit}{bold}{\encodingdefault}{\sfdefault}{bx}{n}

\usepackage[utf8]{inputenc}
\usepackage[T1]{fontenc}
\usepackage{microtype}
\usepackage{hyperref}
\usepackage{url}
\usepackage{booktabs}
\usepackage{amsfonts}
\usepackage{amsmath,amssymb}
\usepackage{graphicx}
\usepackage{multirow}
\usepackage{nicefrac}
\usepackage[table]{xcolor}
\usepackage{array}
\usepackage{caption}
\usepackage{subcaption}
\usepackage{float}
\usepackage{placeins}
\usepackage{tabularx}
\usepackage[most]{tcolorbox}
\usepackage{fontawesome5}
\definecolor{paperlinkpurple}{rgb}{0.28,0.05,0.38}
\hypersetup{
  pdfborder=0 0 0,
  colorlinks=true,
  linkcolor=paperlinkpurple,
  citecolor=paperlinkpurple,
  filecolor=paperlinkpurple,
  urlcolor=paperlinkpurple,
}
\newcolumntype{C}{>{\centering\arraybackslash}X}

\graphicspath{{./}}

\colorlet{tableborder}{black!28}
\colorlet{tableheaderbg}{black!6}
\colorlet{tablebandbg}{black!2}

\newcommand{\paperTableSetup}{%
\arrayrulecolor{tableborder}%
\rowcolors{2}{tablebandbg}{white}%
}

\newcommand{\paperTableHeader}{%
\rowcolor{tableheaderbg}\bfseries
}

\newcommand{\paperTableReset}{%
\rowcolors{2}{}{}%
}
\newtcolorbox{takeaway}{
  enhanced,
  breakable,
  colback=black!3,
  colframe=black!45,
  boxrule=0.4pt,
  arc=2pt,
  left=7pt,
  right=7pt,
  top=4pt,
  bottom=4pt,
  before skip=6pt,
  after skip=6pt
}

\title{Circuit Claims Depend on What Is Extracted\\and How It Is Compared}

\author{%
Yang Sheng$^{1,2,3}$ \qquad Jie Fu$^{3}$\thanks{Corresponding author.}\\
{\normalfont\small $^{1}$Fudan University \quad $^{2}$Shanghai Innovation Institute \quad $^{3}$IQuest Research}\\[-0.15em]
{\normalfont\small Correspondence: \texttt{yann.sheng123@gmail.com; jie.fu@iquestlab.com}}\\[-0.15em]
{\normalfont\small Code: \href{https://github.com/Stepuuu/circuit-extraction-stability}{\faIcon{github}\ \texttt{github.com/Stepuuu/circuit-extraction-stability}}}%
}

\begin{document}

\setcounter{footnote}{1}
\maketitle

\begin{abstract}
Circuit extraction identifies a small set of model components whose presence preserves a target behavior under ablation, and the resulting circuit is often read as the mechanism behind that behavior. We argue that this reading is under-determined: preserving behavior does not single out one circuit, because the claim it supports depends on which circuit is reported and how two circuits are compared. We make this concrete in a synthetic Lean tactic-prediction benchmark---predicting the next step of a proof---where fixed proof rules with randomized surface form let differences between extracted circuits be attributed to these choices rather than to the task. Across checkpoints spanning dense and several weight-sparse levels (most weights constrained to zero) of the same transformer, evaluated on atomic (single-rule) and compositional (multi-rule) proofs, we vary which extracted object is reported (a compact prediction-preserving circuit, a broader graph that also retains the surrounding read, write, and routing structure, or a subgraph extracted with a specified post-ablation loss budget). First, we find low overlap between checkpoints when comparing groups of component-to-component connections, sometimes close to a random baseline. Overlap is much higher when comparing sets of selected attention heads under deterministic selection, but falls sharply when selection among equally scored heads is randomized. Second, the normalized graph-size ranking of conditions that differ in which supervised checkpoint initializes reinforcement learning (RL) agrees between repeated extractions of the compact circuit and between joint and separate query/key representations of each of the other two graphs. For the loss-budgeted graph, query/key ordering agreement persists through the budget sweep. Finally, across those same conditions, the largest exact-match accuracy gains from RL on compositional proofs come with the largest proportion of components in the compositional circuit that lie outside the corresponding atomic circuits. Ablating these additional components in the sparse RL model reduces compositional accuracy while largely preserving accuracy on the corresponding atomic tasks. A circuit-level claim therefore requires specifying not only which circuit is reported, but also how it is extracted and compared; otherwise these unstated choices, not behavior alone, determine the circuit's interpretation. We distill these requirements into a reporting practice for circuit-extraction studies.

\end{abstract}

\begin{figure*}[t]
\centering
\includegraphics[width=\linewidth]{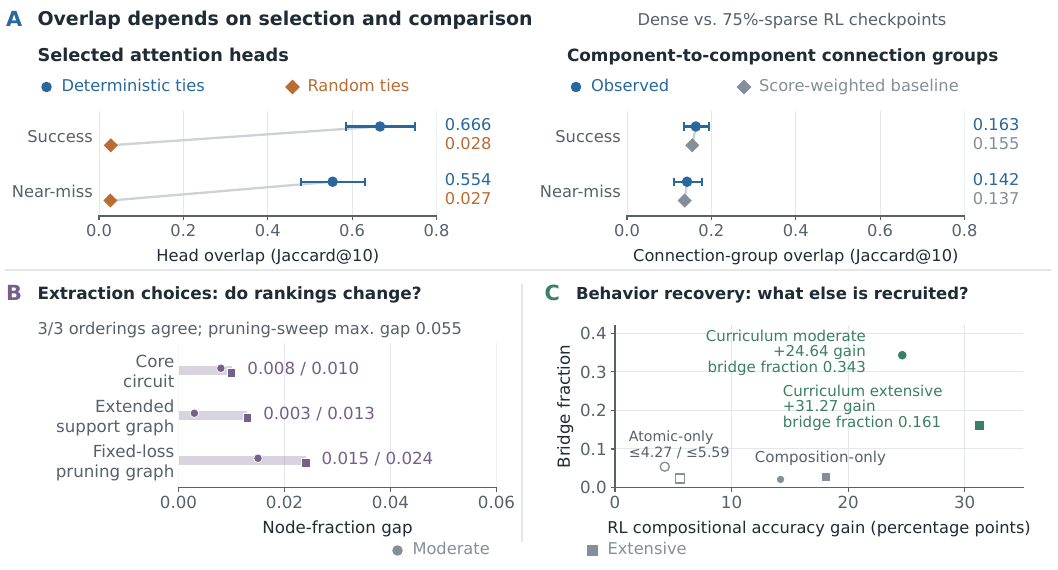}
\vspace{-1.05em}
\caption{Overview of the three findings. \textbf{A}: attention-head and structural edge-bundle overlap at the dense-versus-$75\%$-sparse RL comparison. Circles show observed means; error bars show 95\% bootstrap confidence intervals over 16 matched task--object entries per split. Diamonds show means for random tie handling (heads) or the score-weighted random baseline (connection groups). \textbf{B}: node-fraction gaps for curriculum-initialized RL, comparing repeated core extractions and joint versus separate query/key representations of the other two graphs (pruning loss budget $\tau=0.12$). \textbf{C}: compositional-task exact-match accuracy gain versus core-circuit bridge fraction across RL initialization conditions. Open Atomic-only symbols mark upper bounds on compositional accuracy gain. In B and C, circles and squares denote the moderately and extensively expanded distributions, respectively. Values in B are moderate/extensive.}

\label{fig:paper_overview}
\end{figure*}

\section{Introduction}

Understanding which parts of a neural network implement a behavior is a central goal of mechanistic interpretability \citep{Elhage2021}. One common operationalization is circuit extraction: identifying a small set of components whose presence is enough to preserve or explain a target behavior under ablation \citep{DeCao2022SparseInterventions,Bhaskar2024EdgePruning,Yu2024FunctionalFaithfulness}. Circuit analyses have produced concrete explanations in language models, synthetic algorithmic settings, and sparse feature representations \citep{Wang2023IOI,Conmy2023,Marks2025SparseFeatureCircuits,Dunefsky2024Transcoders}. However, every circuit analysis is shaped by multiple methodological decisions: which extracted graph is reported (a compact prediction-preserving circuit versus a broader graph that also retains surrounding read, write, or routing structure), which ablation or reference distribution defines behavior preservation, whether each attention head is represented by one merged query/key node or by separate query and key nodes, and whether two circuits are compared as structural edge-bundle lists or as attention-head sets. Some of these decisions are discussed explicitly in prior work; others are typically implicit. These choices can change whether two checkpoints appear to share a circuit. For example, in our dense-versus-sparse comparison, the same extracted graphs yield high attention-head overlap under a deterministic top-ten rule, but much lower overlap when membership within tied score groups is randomized (Figure~\ref{fig:paper_overview}A; quantified in Section~\ref{sec:routing_anchor}).

In many natural-language tasks, the effect of these choices is hard to separate from genuine differences in mechanism: the behavior can mix lexical cues, syntax, surface format, and several latent subskills, and the atomic pieces are not specified by construction. A change under another ablation rule, reference distribution, or graph-selection rule may reflect either mechanism or surface cue. In indirect-object identification \citep{Wang2023IOI}, the corrupted reference is itself a design choice: swapping the two names versus altering the sentence template can make different heads appear necessary, with behavior alone unable to say whether the difference is mechanistic or an artifact of what each reference leaves untouched. Recent validation work shows that such choices can change faithfulness scores and reported circuit claims in practice \citep{Miller2024FaithfulnessMetrics,Shi2024CircuitHypothesis,Tigges2024CircuitConsistency}. The question is therefore not only whether a circuit preserves behavior, but how robust that claim is to the choice of which circuit is reported and how circuits are compared.

We study this problem in a synthetic, rule-generated Lean tactic-prediction benchmark. Each input is a short proof state (available hypotheses plus a goal), and the target label is the next tactic, a Lean command that advances the proof. In the examples below, $h_p:p$ means that hypothesis $h_p$ proves proposition $p$ (and declarations such as $p,q,d:\mathrm{Prop}$ declare propositions); $\vdash$ separates the available hypotheses from the goal to prove:
\begin{quote}
\small
\noindent\textbf{Atomic \textsc{AND}}\\[-0.15em]
$p,q,d:\mathrm{Prop};\ h_p:p,\ h_q:q,\ h_d:d \vdash p \land q$\\
Target tactic: \texttt{exact And.intro h\_p h\_q}\\[0.35em]
\noindent\textbf{Compositional \textsc{AND-OR}}\\[-0.15em]
$p,q,r,d:\mathrm{Prop};\ h_p:p,\ h_q:q,\ h_d:d \vdash (p \land q) \lor r$\\
Target tactic: \texttt{exact Or.inl (And.intro h\_p h\_q)}
\end{quote}
The atomic \textsc{AND} example asks the model to prove $p \land q$ from proofs of $p$ and $q$; the target tactic builds that conjunction with \texttt{And.intro}. The compositional \textsc{AND-OR} example asks for $(p \land q) \lor r$: because no proof of $r$ is provided, the target tactic chooses the left side of the disjunction with \texttt{Or.inl} and then uses the same conjunction construction inside it. The premise $h_d:d$ is an unused distractor. The atomic and compositional task structures are known by construction, and surface names, distractors, and sampled examples are randomized while the proof rules remain fixed. This controlled setting lets us vary the reported circuit and the way circuits are compared without confounding from changes in the task family. We vary four factors in this setting: (i) model sparsity, (ii) whether each attention head is represented with a single merged node for query and key or with separate query and key nodes, (iii) the allowed post-ablation loss for the pruning graph, and (iv) which supervised checkpoint initializes reinforcement learning (RL).

Figure~\ref{fig:paper_overview} summarizes three related findings, and Table~\ref{tab:main_claim_map} maps each to its evidence. The first examines how comparison choices determine circuit-overlap claims; the other two examine graph-size orderings and the relation between circuit structure and behavior. (1) High overlap between attention-head sets under deterministic selection is sensitive to how tied scores are resolved. (2) Within each graph, RL initialization conditions have consistent normalized-size orderings between repeated core extractions or between joint and separate query/key representations. (3) The conditions with the largest exact-match accuracy gains from RL on compositional proofs have the largest bridge fractions. Full bridge-node ablation in the sparse RL model reduces compositional accuracy while largely preserving accuracy on the corresponding atomic tasks.

\section{Related Work}

Mechanistic interpretability has developed residual-stream circuit formalisms \citep{Elhage2021}, targeted case studies \citep{Wang2023IOI}, automated discovery pipelines \citep{Conmy2023}, path patching \citep{GoldowskyDill2023PathPatching}, scalable attribution localization \citep{Kramar2024AtPStar}, and learned feature circuits \citep{Dunefsky2024Transcoders}. \citet{Hanna2024Faithfulness} introduce EAP-IG to improve circuit faithfulness and show that high circuit overlap can coexist with differences in faithfulness. \citet{Davies2023Desiderata} use causal desiderata to localize variable-binding components. Closest to the question here, causal-abstraction work shows that the same network can admit multiple valid higher-level descriptions \citep{Geiger2023CausalAbstraction}, faithfulness evaluations show sensitivity to validation and extraction choice \citep{Miller2024FaithfulnessMetrics,Shi2024CircuitHypothesis}, and circuit-stability work finds that algorithms and component types can persist across training and scale even when individual attention-head implementations change \citep{Tigges2024CircuitConsistency}. Recent work argues that a single task in an LLM can be supported by multiple structurally distinct faithful circuits or sheaves, using an overlap penalty to discover low-overlap solutions \citep{Chen2026AllCircuitsRome}. Fine-tuning studies provide a related positive case in which entity-tracking circuits persist across training stages \citep{Prakash2024FineTuningMechanisms}. \citet{Meloux2025DeadSalmons} frame interpretability as statistical inference, emphasizing identifiability and meaningful alternative hypotheses. Together, these studies examine the sensitivity of faithfulness metrics and circuit estimates to methodological choices \citep{Miller2024FaithfulnessMetrics,Shi2024CircuitHypothesis,Meloux2025Variance}, the non-identifiability of circuits that equally preserve a target behavior \citep{Geiger2023CausalAbstraction,Chen2026AllCircuitsRome,Meloux2025Identifiable}, disagreement among explanations for one model \citep{Krishna2022Disagreement}, and, in contrast, the consistency of coarser circuit structure across training and scale \citep{Tigges2024CircuitConsistency}. Those studies ask whether mechanisms persist across training, scale, or fine-tuning, or whether alternative faithful mechanisms can be found; our work asks a different question: when benchmark and checkpoints are held fixed, which claims about the extracted circuit---for example, whether two checkpoints share a circuit, or whether one RL condition selects more bridge components---change when the extracted graph changes, when query/key attention support is merged or separated, when the post-ablation loss threshold changes, or when the comparison level changes.

Our experiments also draw on sparse-model and pruning-based interpretability. Differentiable masking and pruning methods can isolate small causal subsets of model components \citep{DeCao2022SparseInterventions,Bhaskar2024EdgePruning,Yu2024FunctionalFaithfulness}, while benchmark environments such as Tracr, InterpBench, and the Mechanistic Interpretability Benchmark (MIB) make exact recovery and faithfulness easier to test \citep{Lindner2023Tracr,Gupta2024InterpBench,Mueller2025MIB}. Recent sparse-model and sparse-feature-circuit work suggests that sparsity in model structure or learned representations can make mechanisms more inspectable \citep{Gao2025,Draye2025,Marks2025SparseFeatureCircuits}. For the compositional comparison, \citet{Mondorf2025CircuitCompositions} compare circuits for modular string-edit operations and test reuse/composition through circuit set operations; by contrast, our experiments vary which graph is extracted and whether query and key sides of attention are merged or separated while checkpoints stay fixed.

Controlled compositional benchmarks motivate the benchmark design. Synthetic compositional-generalization benchmarks test whether learned behaviors transfer across rule-composed inputs \citep{LakeBaroni2018,Keysers2020,KimLinzen2020,Ruis2020,Hupkes2020,Clark2020,Tafjord2021,Yang2023}, and behavioral testing work shows how surface-form controls can change evaluation conclusions \citep{Ribeiro2020,WuManningPotts2023}. Our synthetic Lean benchmark borrows this controlled-generation principle but applies it to circuit extraction rather than behavioral evaluation: the proof-state rules and the atomic-versus-compositional decomposition are fixed and known by construction, while surface form and reasoning depth vary. This known decomposition is what lets us attribute differences between extracted circuits to the extraction and comparison choices rather than to uncontrolled changes in the task.

\section{Experimental Setup}
\label{sec:setup}

\subsection{Benchmark and Training Distributions}

We construct a benchmark of nine synthetic Lean tactic-prediction tasks: four atomic and five compositional (Table~\ref{tab:benchmark_tasks}). Each example is generated dynamically from fixed propositional proof rules, with randomized proposition and hypothesis names plus distractor premises, and the model predicts the first tactic that advances the proof state. As a concrete instance, an atomic example might ask the model to prove $p \land q$ from $h_p:p$, $h_q:q$, and a distractor $h_d:d$, where the target tactic is \texttt{exact And.intro h\_p h\_q}; a compositional example might ask the model to prove $(p \land q) \lor r$ from the same three hypotheses, where the target tactic is \texttt{exact Or.inl (And.intro h\_p h\_q)}. The compositional goal introduces a new proposition $r$ for which no hypothesis is given, so the proof must use the left disjunct.

The synthetic Lean family is chosen for the three properties that a controlled circuit comparison requires. First, the proof-state generation rules are fixed, so atomic and compositional structures vary independently of surface form. Second, randomized names, distractors, and dynamically generated examples reduce fixed-string memorization, though the boundary tests below show that shared format cues can still matter. Third, because examples are dynamically generated rather than drawn from a fixed test set, a single test draw could over- or under-state accuracy by chance; therefore, we report behavioral results on two independently sampled test sets of $100$ examples per task, labeled Suite~1 and Suite~2.

To obtain supervised checkpoints that differ only in the breadth of their training distribution, we train on a narrow baseline distribution and two expanded distributions that share the same proof-state generation rules while varying surface form. The moderately expanded distribution adds distractor-heavy, goal-first, and long-name variants; the extensively expanded distribution additionally includes compact and relabeled formats. These checkpoints become the initialization conditions for the RL comparison, and because only the surface form changes while the proof-state rules stay fixed, differences across them---and across the circuits extracted from them---reflect training-distribution breadth rather than a change in the task.

\begin{figure}[tbp]
\centering
\begin{subfigure}[t]{0.49\linewidth}
    \centering
    \includegraphics[width=0.66\linewidth]{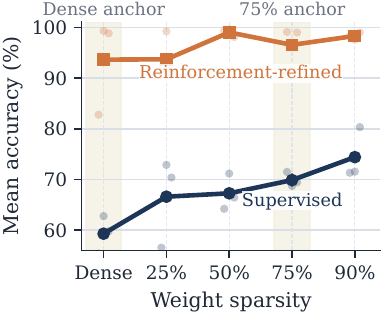}
    \caption{Baseline accuracy across sparsity levels.}
    \label{fig:baseline_behavior}
\end{subfigure}\hfill
\begin{subfigure}[t]{0.49\linewidth}
    \centering
    \includegraphics[width=0.66\linewidth]{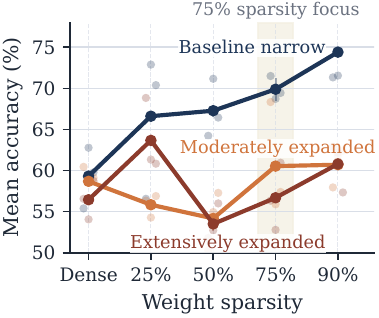}
    \caption{Supervised accuracy after distribution expansion.}
    \label{fig:expanded_supervised}
\end{subfigure}
\caption{Behavioral accuracy across sparsity levels. \textbf{A}: the baseline training distribution across dense and four weight-sparse regimes; the highest-accuracy checkpoint and the checkpoint used for detailed graph extraction need not coincide. \textbf{B}: at the same sparsity points, expanded training distributions reduce mean supervised accuracy across the two evaluation suites, especially at the $75\%$ weight-sparse graph-extraction checkpoint.}
\label{fig:supervised_behavior_maps}
\end{figure}

This generator design combines controlled compositional generation with held-out surface variation \citep{LakeBaroni2018,Keysers2020,KimLinzen2020,Ruis2020,Hupkes2020,Ribeiro2020,WuManningPotts2023} and the use of controlled benchmarks for interpretability evaluation \citep{Lindner2023Tracr,Gupta2024InterpBench,Mueller2025MIB}. We instantiate these ideas in a proof-oriented setting related to natural-language proof generation and Lean theorem proving \citep{Tafjord2021,Yang2023}. We use Lean tactics as supervision targets because the label is an exact tactic string, giving an unambiguous supervision signal and a hard exact-match metric for both behavioral evaluation and ablation-based circuit selection. Because examples are generated dynamically, supervised data volume is specified by token budget rather than by a fixed corpus size: each supervised run consumes $5$B tokens drawn from one of these distributions, and each RL run uses $500$ composition-only prompts per task with $8$ generations per prompt over $10$ epochs (full settings in Table~\ref{tab:training_recipes}).

\subsection{Model Families and Training Stages}

We train weight-sparse models for two reasons: weight sparsity yields smaller, more inspectable circuits, which makes circuit extraction and comparison tractable \citep{Gao2025}, and the level of sparsity is itself one of the factors whose effect on circuit claims we measure. Our weight-sparse training follows the top-$k$ convention of that work: $x\%$ weight-sparse means $x\%$ of weights are constrained to zero during training. The main experimental grid uses an 8-layer transformer architecture ($d_{\mathrm{model}}=2048$, $n_{\mathrm{head}}=128$, $d_{\mathrm{head}}=16$, $d_{\mathrm{mlp}}=8192$, context length $128$, $\sim\!415.7$M parameters), trained on our Lean tactic-prediction benchmark at five sparsity levels: dense, $25\%$, $50\%$, $75\%$, and $90\%$ weight-sparse. This range is large enough to learn the task family across dense, weight-sparse, supervised, and reinforcement learning (RL) settings, yet small enough for repeated extraction, ablation, and seed/sparsity comparisons. Three independent seeds are trained at every sparsity level in this grid. We also compare four-layer SFT models to test sensitivity to depth. SFT uses RMSNorm; GRPO loads the SFT weights into LayerNorm models with the same dimensions.

Our experiments include the baseline distribution and a grid crossing the two expanded distributions with three reinforcement-learning (RL) initialization conditions. The three RL initialization conditions in each expanded-distribution family differ only in their upstream supervised checkpoint: \emph{Curriculum} starts from the full curriculum-trained checkpoint, \emph{Atomic-only} starts from a checkpoint trained on atomic tasks only, and \emph{Composition-only} starts from a checkpoint trained on compositional tasks only. The prompt mix used for Group Relative Policy Optimization (GRPO; \citealp{Shao2024GRPO}) remains composition-only across all three initialization conditions; the condition labels refer to upstream checkpoint identity, not to the RL prompt distribution. Detailed evidence coverage is in Table~\ref{tab:evidence_scope}.

Behavioral comparisons use all five sparsity levels. The main overlap comparison and representative graph visualizations in the baseline distribution use dense and $75\%$ weight-sparse checkpoints; the expanded-distribution graph-size and bridge comparisons aggregate over five sparsity levels and three seeds. Dense serves as the unpruned reference; $75\%$ weight sparsity is a pre-specified graph-extraction point, chosen to balance graph readability, object size, and reuse of one dense-vs-weight-sparse comparison in the overlap analysis and representative graph visualizations. Figure~\ref{fig:supervised_behavior_maps} situates these checkpoints within the full five-sparsity behavioral sweep: $75\%$ maintains strong supervised and RL accuracy in the baseline family, while expanded training distributions reduce mean supervised accuracy across the two evaluation suites at the same sparsity points. Apart from this checkpoint choice, training and checkpoint-selection protocols are held fixed across the compared families: supervised runs share the 5B-token budget and optimization settings, and the narrow baseline RL family and all expanded-distribution RL initialization conditions share the same GRPO refinement and model-selection rule \citep{Shao2024GRPO}. Table~\ref{tab:training_recipes} gives the shared training protocol. We focus on RL checkpoints because their higher task accuracy provides more prompts solved by both endpoints.

\subsection{Circuit Objects, Query/Key Support, and Comparisons}
\label{sec:setup_objects}

We report three extracted circuit objects: the \emph{core circuit}, a compact prediction-preserving subgraph; the \emph{extended support graph}, a wider graph that also keeps surrounding read, write, and routing components; and the \emph{fixed-loss pruning graph}, a subgraph selected from a learned ranking under a specified post-ablation loss budget. They differ in which components they keep and how they are selected, so that the same sensitivity question can be checked across multiple extracted graphs. The objects combine two extraction criteria from prior work with a broader reporting view, using node types adapted from the residual-stream circuit framing of \citet{Elhage2021} and the fine-grained node families used in \citet{Gao2025}. The core circuit selects attention-output channels and MLP neurons. The extended support and fixed-loss pruning graphs use a broader component representation that includes attention read, query/key, value, and write components, as well as MLP read, neuron, and write components. Throughout, the \emph{support} of a node denotes the set of upstream components in this extraction graph whose retained activations feed that node---for example, the residual-stream reads that feed an attention head's query, key, or value channel at a given token position, or the MLP and attention writes that carry information back into the residual stream. An extracted circuit is the pair of selected nodes and the support edges linking them. Below, a change in how the circuit is extracted means changing one of selection rule, criterion, or query/key node choice while holding the checkpoint, task set, and relevant loss threshold fixed.

\noindent\textbf{Core circuit.} The \emph{core circuit} is a compact extracted subgraph selected because, after the other candidate components are ablated, the remaining components still preserve the target tactic prediction. This object instantiates the prediction-preserving learned-mask extraction line of \citet{DeCao2022SparseInterventions,Bhaskar2024EdgePruning,Yu2024FunctionalFaithfulness} over attention-output channels and MLP neurons. We adapt this line in two places: (i) the primary selection criterion is exact-match target-tactic accuracy after ablation, with the masked cross-entropy plus $\lambda_{\ell_0}$ penalty used only as a tiebreaker; (ii) the mask objective is supervised on Lean tactic tokens. Concrete hyperparameters are in Table~\ref{tab:extraction_recipes}.

\noindent\textbf{Extended support graph.} The \emph{extended support graph} uses the same prediction-preserving learned-mask approach as the core circuit \citep{DeCao2022SparseInterventions,Bhaskar2024EdgePruning,Yu2024FunctionalFaithfulness}, with a broader node vocabulary that includes read, write, and routing components around the same prediction-preserving behavior. This wider graph is a deliberately broader reporting view rather than a separate extraction criterion; it tests whether claims that hold for the compact core also hold when surrounding support structure is included.

\noindent\textbf{Fixed-loss pruning graph.} The \emph{loss-constrained pruning subgraph} instantiates the loss-budget circuit-selection line of \citet{Conmy2023,Bhaskar2024EdgePruning}, and is the closest analogue in our setup of the task-specific pruning step in \citet{Gao2025}. After optimizing a continuous top-$k$ pruning-score objective, the procedure evaluates circuits $C_k$ retaining the top $k$ components in the fitted ranking. Among the evaluated sizes $\mathcal{K}_{\mathrm{tested}}$, it selects $k^\star=\min\{k\in\mathcal{K}_{\mathrm{tested}}:\mathcal{L}_{\mathrm{abl}}(C_k)\le\tau\}$ when this set is nonempty. Appendix~\ref{app:extraction} specifies the search and fallback rules when no evaluated circuit meets the budget. We adapt this line with a two-stage continuous-then-discrete top-$k$ selection for our Lean tactic-prediction loss. For readability, this object is called the fixed-loss pruning graph below. The main coupled-vs-factorized comparison uses $\tau=0.12$; a loss-budget sweep applies the fitted pruning-score ranking at additional $\tau$ values to test whether the ordering conclusion depends on this pruning threshold.

For the extended support and fixed-loss pruning graphs, we compare two ways to represent query-side and key-side support for attention heads. Prior circuit-extraction and localization methods choose a graph granularity before searching or patching circuits \citep{Conmy2023,GoldowskyDill2023PathPatching,Bhaskar2024EdgePruning}; here we compare coupled and factorized query/key representations using the object-specific extraction recipes in Table~\ref{tab:extraction_recipes}. For an attention head $H$ at layer $\ell$, its \emph{query-side support} is the set of upstream components feeding $H$'s query projection at the target token position, and its \emph{key-side support} is the analogous set feeding $H$'s key projection (typically read from earlier positions). \emph{Coupled Q/K}, the merged query/key representation, uses a shared \texttt{attn\_qk} node family for query-side and key-side support. \emph{Factorized Q/K}, the separated representation, keeps them as separate \texttt{attn\_q} and \texttt{attn\_k} node families, while keeping the read, value, write, MLP-read, MLP-neuron, and MLP-write node families explicit. The set of possible nodes and edges can therefore change even when the same coarse read-to-write route is present. For the extended support graph, the learned-mask optimization recipe and selection rule are held fixed across the coupled and factorized versions. The core circuit has no separate query/key nodes; its comparison repeats extraction with the same node representation. Table~\ref{tab:extraction_recipes} summarizes the concrete extraction recipes, selection criteria, and loss or accuracy criteria.

\paragraph{How claims are compared.}
Each comparison tracks whether the qualitative takeaway stays the same across repeated extraction, query/key representations, or pruning loss budgets: whether the ordering or sign stays the same, for example, whether one RL initialization condition still has the larger node fraction, or whether overlap between attention-head sets still exceeds structural overlap.

\paragraph{Comparison rule.}
Comparison is object-specific. The three circuit objects differ in node types, edge types, and extraction objectives by design, so claims are compared separately for each object.

\subsection{Metrics and Statistics}
\label{sec:setup_metrics}

Four classes of metrics support the comparisons.

\noindent\textbf{Behavior.} Behavioral accuracy reports overall accuracy on both evaluation suites together with atomic-task and compositional-task means, following the exact-match accuracy used in compositional-generalization and proof-prediction benchmarks \citep{LakeBaroni2018,Hupkes2020,Yang2023}. When a single summary is useful, the mean accuracy is
\[
\mathrm{Acc}_{\mathrm{mean}}=\tfrac{1}{2}\bigl(\mathrm{Acc}_{\mathrm{suite1}}+\mathrm{Acc}_{\mathrm{suite2}}\bigr).
\]
\noindent\textbf{Structural overlap and attention-head overlap.} We use the standard Jaccard index for set overlap \citep{Jaccard1901}, which recent interpretability work also uses to compare discovered circuits \citep{Tigges2024CircuitConsistency,Meloux2025Variance}: the size of the intersection divided by the size of the union, so $0$ means no shared selected items and $1$ means identical selected sets. In our Jaccard@10 summaries, each set is first restricted to the top $10$ selected entries for that metric. For relation type $r\in\{\mathrm{structural},\mathrm{semantic},\mathrm{routing}\}$,
\[
J_k^{(r)}(A,B)=\frac{\lvert \mathrm{Top}_k^{(r)}(A)\cap \mathrm{Top}_k^{(r)}(B)\rvert}{\lvert \mathrm{Top}_k^{(r)}(A)\cup \mathrm{Top}_k^{(r)}(B)\rvert},
\]
with $k=10$ throughout. For structural comparisons, we group support connections by layer, source and target component type, and their membership in the atomic and compositional circuits. We call these groups \emph{structural edge bundles}. A membership category records how a selected component relates to the atomic circuits for the same compositional task: for example, whether it is shared with an atomic circuit, reused from an atomic circuit, or selected mainly in the compositional circuit. For example, \texttt{L0::shared->shared::mlp\_neuron->mlp\_write} groups connections with those structural attributes. We sum the constituent connection scores within each group and compare the top-$k$ groups by Jaccard. We call this measure structural edge-bundle overlap, or structural overlap for short. Semantic overlap coarsens the source and target membership categories into shared-with-atomic, reused-atomic, compositional-only, and other classes before computing Jaccard; it is numerically indistinguishable from structural overlap in these comparisons.

For the head comparison, we count routing connections per attention head and select the heads with the largest counts; these selected heads form the \emph{routing-head set}. Deterministic selection resolves equal scores by head enumeration order. Section~\ref{sec:routing_anchor} compares this rule with random selection among equally scored heads. Individual weight-edge overlap, defined in Appendix~\ref{app:weight_edge_audit}, compares entries of weight matrices.

The structural and routing-head comparisons use the same pre-specified dense and $75\%$ weight-sparse RL checkpoints, matching entries by task, extracted object, and generation seed. The success split uses the generation-seed selections associated with successful predictions at both endpoints. The near-miss split uses non-success selections closest to the correct tactic output under a tactic-aware token and exact-match score (Appendix~\ref{app:extraction}). Each split contains 16 matched task--object entries with scored bundles at both endpoints.

\noindent\textbf{Node fractions and extraction gaps.} Compositional-task node fraction is the mean selected-node count divided by the full object-specific node universe over the five compositional tasks, with the universe defined by the given object and query/key support choice. The coupled-vs-factorized Q/K fraction gap compares the two versions of the same object within the same family-condition cell, $\Delta_{\mathrm{frac}}=\lvert f_{\mathrm{coup}}-f_{\mathrm{fact}}\rvert$. The core-circuit gap instead compares two runs of the same extractor. Each fraction uses its object's own node universe, so gaps are within-object summaries rather than common-scale effect sizes across objects. Motivated by circuit-set comparisons of compositional circuits \citep{Mondorf2025CircuitCompositions}, we define bridge fraction per task as $\mathrm{BridgeFrac}(t)=\left|C_t \setminus \bigcup_i C_t^{(i)}\right|/\left|C_t\right|$, where $C_t$ is the circuit extracted from the composite task and $C_t^{(i)}$ are the corresponding atomic-task circuits. We call the nodes in $C_t \setminus \bigcup_i C_t^{(i)}$ bridge nodes. The reported bridge fraction is the mean of $\mathrm{BridgeFrac}(t)$ over the five compositional tasks.

\noindent\textbf{Aggregation, uncertainty, and controls.} Full sparsity sweeps report means across three seeds and visualize per-seed scatter. For the primary structural edge-bundle comparison, we compute 95\% bootstrap confidence intervals over the 16 matched task--object entries per split \citep{Efron1979,Meloux2025Variance}. Additional audits report their own coverage and randomization protocol; a failure to reject a null is not an equivalence test. The RL initialization-condition plots aggregate means over five sparsity levels and three seeds per family-condition cell. We use three negative-control conditions, aggregated over the same five compositional tasks: matched atomic keeps the union of the atomic-task circuits that define the compositional task, mismatched atomic keeps a low-overlap atomic union from the same checkpoint, and random averages size-, layer-, and layer-family-matched random controls.

\section{Results}
\label{sec:results}

We first examine how comparison choices affect claims that two checkpoints share a circuit. We then test whether graph-size orderings across RL initialization conditions persist under alternative extractions, and relate bridge structure to compositional behavior. Table~\ref{tab:main_claim_map} maps these findings to their evidence.

\begin{table}[t]
\caption{Summary of the main findings. Metric definitions are given in Section~\ref{sec:setup_metrics}; full numerical values, confidence intervals, random-baseline checks, and controls are reported in the cited figures and tables. Node-fraction gaps are within-object summaries; $\tau=0.12$ is the main loss-budget point.}
\label{tab:main_claim_map}
\centering
\scriptsize
\setlength{\tabcolsep}{2.4pt}
\renewcommand{\arraystretch}{1.12}
\paperTableSetup
\begin{tabularx}{\linewidth}{>{\raggedright\arraybackslash}p{0.18\linewidth} >{\raggedright\arraybackslash}p{0.32\linewidth} >{\raggedright\arraybackslash}p{0.13\linewidth} >{\raggedright\arraybackslash}X}
\toprule
\paperTableHeader Finding & Key result & Evidence & Comparison \\
\midrule
Overlap depends on selection and comparison & Routing-head Jaccard@10 (success/near-miss splits) is $0.666/0.554$ under deterministic selection and $0.028/0.027$ with random tie handling. Structural overlap is $0.163/0.142$, close to score-weighted baselines of $0.155/0.137$. & Figure~\ref{fig:edge_overlap}; Tables~\ref{tab:edge_overlap}, \ref{tab:edge_overlap_random}, \ref{tab:exact_edge_supplement}; Appendix~\ref{app:rebuttal-ranking} & Dense vs.\ $75\%$ weight-sparse RL checkpoints. \\
\addlinespace[3pt]
Graph-size orderings agree across extraction variants & Node-fraction gaps are at most $0.024$ at $\tau=0.12$; all pairwise RL-condition orderings agree ($3/3$). Q/K ordering agreement is preserved throughout the pruning-budget sweep. & Figure~\ref{fig:object_formalization}; Tables~\ref{tab:mechanism_stage}, \ref{tab:mechanism_stage_ordering}, \ref{tab:mechanism_stage_bootstrap}, \ref{tab:tau_sweep_summary} & Repeated core extraction; coupled vs.\ factorized Q/K for the support and pruning graphs; fixed-score pruning-budget sweep. \\
\addlinespace[3pt]
Compositional RL gains track bridge fraction & Curriculum RL improves compositional accuracy by $+24.64/+31.27$ percentage points, with core bridge fractions of $0.343/0.161$ (moderate/extensive); other RL conditions have bridge fractions $\le 0.054$. Full bridge ablation reduces compositional accuracy while largely preserving matched atomic accuracy. & Figure~\ref{fig:expanded_rl_behavior}; Tables~\ref{tab:expanded_rl}, \ref{tab:mechanism_stage}; Appendix~\ref{app:rebuttal-bridge} & Across RL initialization conditions; full bridge-node ablation in the sparse RL model. \\
\bottomrule
\end{tabularx}
\paperTableReset
\end{table}

\begin{figure}[tbp]
\vspace{0.6em}
\centering
\includegraphics[width=0.88\linewidth]{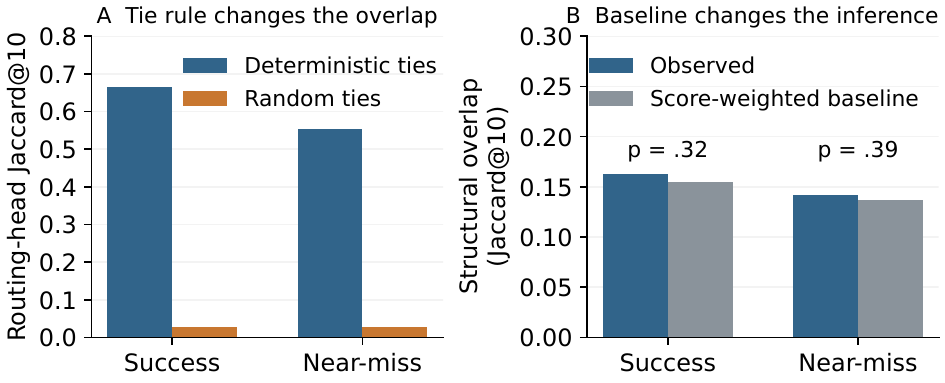}
\caption{Overlap under alternative comparison rules at the dense--75\%-sparse RL pair. Left: attention-head Jaccard@10 under deterministic selection and random selection within tied score groups. Right: overlap between structural edge bundles (Jaccard@10), compared with a score-weighted random baseline. Bars show mean overlap; deterministic-ranking bootstrap intervals are reported in Table~\ref{tab:edge_overlap}.}

\label{fig:edge_overlap}
\end{figure}

\subsection{Overlap Conclusions Depend on Selection and Comparison}
\label{sec:routing_anchor}

Under deterministic selection, attention-head overlap is much higher than overlap between structural edge bundles (Figure~\ref{fig:edge_overlap}). For the dense--$75\%$-sparse RL comparison, attention-head Jaccard@10 is $0.666$ on the success split and $0.554$ on the near-miss split (Table~\ref{tab:edge_overlap}). However, many attention heads receive the same routing score. Randomizing selection among these equally scored heads reduces overlap to approximately $0.028$ and $0.027$, respectively, with the extracted graphs and scores held fixed. The comparisons at other sparsities and across seeds use the same deterministic selection rule (Table~\ref{tab:exact_edge_supplement}). These results show that the high attention-head overlap is sensitive to selection within tied score groups.

\paragraph{Structural overlap and random baselines.}
At the dense--$75\%$-sparse RL comparison, structural Jaccard@10 is $0.163/0.142$ for the success/near-miss splits, close to score-weighted random baselines of $0.155/0.137$ ($p=0.32/0.39$). This baseline samples connection groups using their scores at each checkpoint. For the success split at the other sparsity levels, structural overlap exceeds the corresponding random-baseline means (Table~\ref{tab:additional-overlap-nulls}). Appendix~\ref{app:rebuttal-ranking} gives the sampling rules and the comparison with uniform random sampling.

\paragraph{Repeated extraction and additional model comparisons.}
Repeated extraction from the same RL checkpoint gives structural Jaccard@10 of $0.78$--$0.80$. At $75\%$ sparsity, the size-matched random baseline is also high ($0.75$--$0.77$), while agreement between the rankings is higher than its random baseline. These comparisons assess extraction repeatability within a checkpoint, separately from circuit sharing across checkpoints. Additional SFT and four-layer comparisons give structural overlap close to their size-matched baselines. Appendix~\ref{app:rebuttal_audits} reports the extraction settings, baselines, and model comparisons.

\paragraph{An independent weight-level attribution analysis.}
We adapt the path-integration approach of EAP-IG~\citep{Hanna2024Faithfulness} to score individual weight connections and aggregate their scores by attention head. In this analysis, the top-ten cutoffs contain no ties. Mean attention-head Jaccard@10 is $0.00585$, close to the uniform random baseline of $0.00534$; individual weight-edge overlap and its sampled random baseline are both zero. Ablating the highest-ranked weights increases loss more than the mean of matched random controls in 16 of 18 task--checkpoint cases. Appendix~\ref{app:weight_edge_audit} gives the scoring procedure, checkpoint configuration, and intervention results.

\begin{takeaway}
\textbf{Takeaway.} For the dense and $75\%$ weight-sparse checkpoints, structural overlap is low, whereas attention-head overlap is much higher under deterministic selection but falls sharply when selection among equally scored heads is randomized.
\end{takeaway}

\begin{figure}[htbp]
\centering
\includegraphics[width=0.86\linewidth]{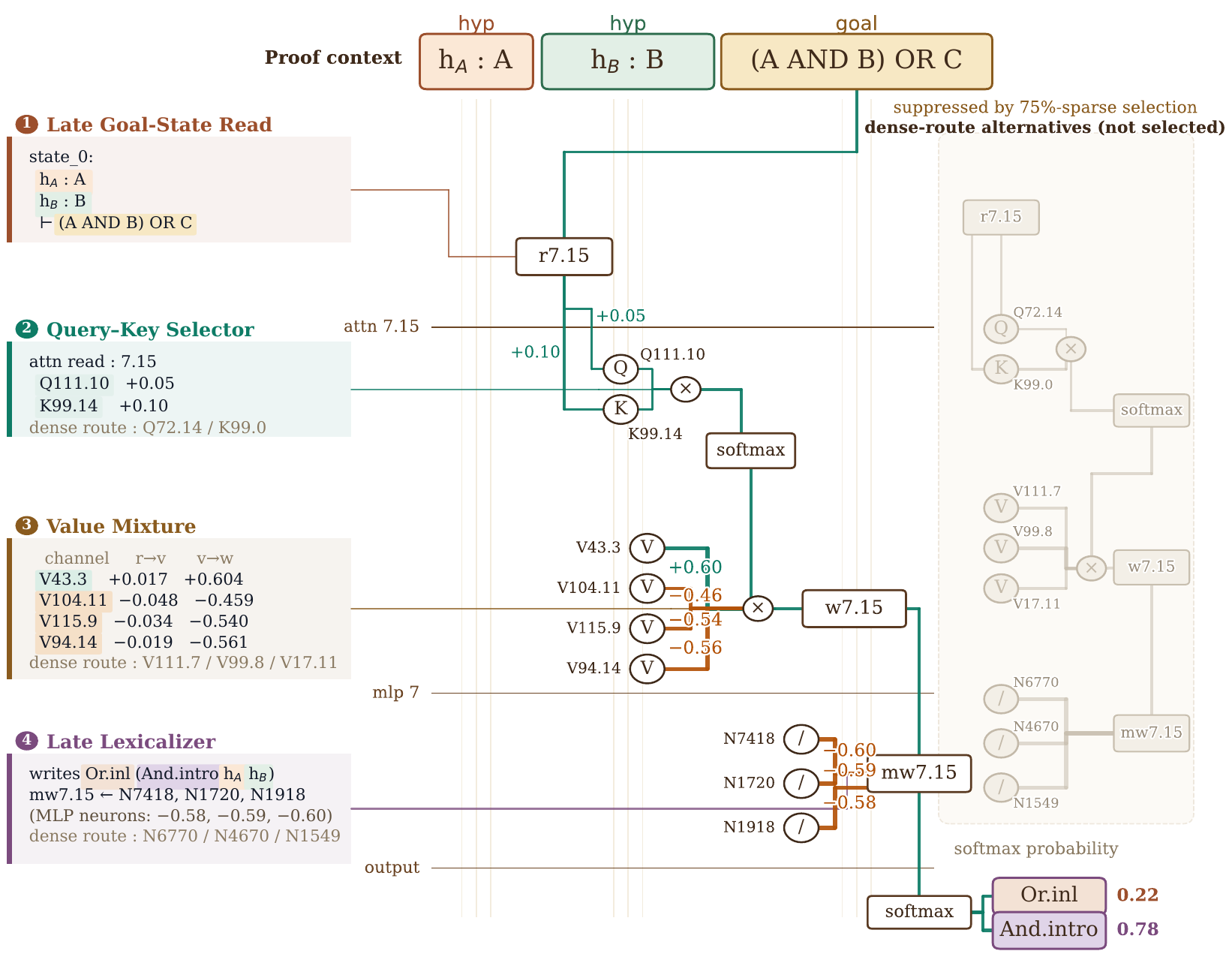}
\caption{Representative $75\%$ weight-sparse RL case for \textsc{AND-OR}. Arrows denote the direction of information flow. The recurring pattern is a coarse read-to-write path: a late-position read $\rightarrow$ a compact query/key selector (the attention pattern that decides what to attend to) $\rightarrow$ a small group of value-vector components carrying signed contributions $\rightarrow$ the attention write back into the residual stream $\rightarrow$ downstream MLP components that raise the target tactic-token logits. The paired dense and weight-sparse examples follow this functional path with different selected components.}
\label{fig:local_mechanistic_story}
\end{figure}

Figure~\ref{fig:local_mechanistic_story} shows a representative $75\%$ weight-sparse RL \textsc{AND-OR} success case from this paired comparison, with the paired dense/weight-sparse node-level case in Figure~\ref{fig:node_level_routing}. The shared structure is a recurring read/write path: a compact query/key selector routes a small value-vector component group into \texttt{attn\_write:7,15}, and downstream MLP components increase the final tactic-token logits for \texttt{Or.inl} and \texttt{And.intro}. Aggregate evidence comes from Figure~\ref{fig:edge_overlap} and the supplement.

\subsection{Graph-Size Orderings Agree Across Extraction Variants}
\label{sec:setup_dependence}

The second comparison holds the checkpoint condition fixed and varies the extraction choice: repeated extraction for the core circuit, coupled versus factorized query/key representations for the extended support and fixed-loss pruning graphs, and the post-ablation loss budget for the fixed-loss pruning graph. Each graph gives a consistent ordering of RL initialization conditions by normalized size across its extraction variants. For the core circuit, the curriculum-initialized conditions achieve the highest compositional accuracy while selecting the smallest node fractions (Tables~\ref{tab:expanded_rl} and~\ref{tab:mechanism_stage}).

Holding the loss budget fixed at $\tau=0.12$, the coupled-vs-factorized query/key gap stays small for every family-condition cell. For curriculum initialization, mean absolute Q/K gaps in the moderately and extensively expanded distributions are $0.003$ and $0.013$ for the extended support graph, and $0.015$ and $0.024$ for fixed-loss pruning. Core fractions differ by $0.008$ and $0.010$ between extraction runs (Table~\ref{tab:mechanism_stage}). Across these within-object summaries, gaps are at most $0.024$.

\begin{figure*}[t]
\centering
\includegraphics[width=\textwidth]{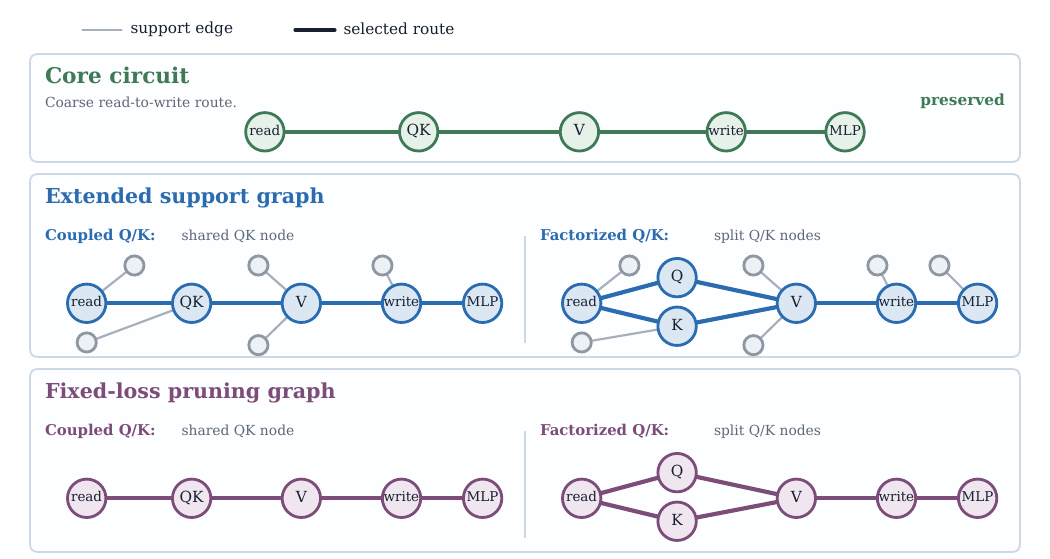}
\caption{Representative extraction under coupled (merged) versus factorized (separate) query/key support at the main loss budget on an RL checkpoint at $75\%$ weight sparsity. This schematic illustrates the object-level pattern at one cell; the aggregate selected-node fractions and RL-condition orderings in Tables~\ref{tab:mechanism_stage}--\ref{tab:mechanism_stage_bootstrap} confirm that the orderings are preserved at this pruning threshold. Table~\ref{tab:tau_sweep_summary} extends this check across the tested range of loss budgets, with $\tau=0.12$ as the main reporting point. Figure~\ref{fig:object_formalization_graph_matrix} gives a more detailed view of this example, grouping selected components by layer and component type.}
\label{fig:object_formalization}
\end{figure*}

These small gaps do not change the RL-condition orderings. After averaging within each family-condition cell over five sparsity levels and three seeds, the two extraction runs agree on all three pairwise condition rankings for every object in both expanded distributions (Table~\ref{tab:mechanism_stage_ordering}). A bootstrap over sparsity-seed units gives the same qualitative result: the observed agreement is $3/3$ for every object and family, with the largest pairwise flip probability ($0.232$) appearing in the extensively expanded fixed-loss pruning comparison between the curriculum-initialized and composition-only conditions (Table~\ref{tab:mechanism_stage_bootstrap}). Figure~\ref{fig:object_formalization} illustrates one representative case; Figure~\ref{fig:object_formalization_graph_matrix} gives the projected graph counterpart for that exemplar.

We hold the fitted pruning-score ranking fixed and sweep $\tau \in \{0.04, 0.08, 0.12, 0.16, 0.20\}$. For the fixed-loss pruning graph, the coupled-vs-factorized curriculum gap stays modest across the full range (at most $0.055$ under the family-level aggregation used for Table~\ref{tab:tau_sweep_summary}), and coupled and factorized Q/K agree on all three pairwise RL-condition comparisons in both families. The condition ordering changes between $\tau=0.04$ and $0.08$; agreement between the two Q/K representations is maintained throughout the sweep (Table~\ref{tab:tau_sweep_summary}).

\begin{takeaway}
\textbf{Takeaway.} RL-condition size orderings agree between repeated core extractions and between Q/K representations of each richer graph. For fixed-loss pruning, the Q/K agreement persists through the budget sweep. These are distinct checks: the core extractor does not split query and key components.
\end{takeaway}

\subsection{Bridge Structure and Its Contribution to Compositional Behavior}
\label{sec:scope_stress_tests}

We next connect core-circuit structure to compositional behavior. We find that curriculum-initialized RL produces both the largest compositional-task gains and the largest core-circuit bridge fraction---the share of compositional-task circuit nodes that lie outside the union of the corresponding atomic-task circuits.

\begin{figure}[t]
\centering
\begin{subfigure}[t]{0.32\linewidth}
    \centering
    \includegraphics[width=0.98\linewidth]{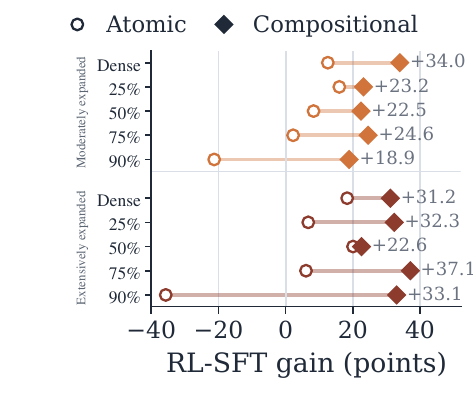}
    \caption{Compositional gains.}
    \label{fig:expanded_rl_atomic_comp}
\end{subfigure}\hfill
\begin{subfigure}[t]{0.31\linewidth}
    \centering
    \includegraphics[width=0.98\linewidth]{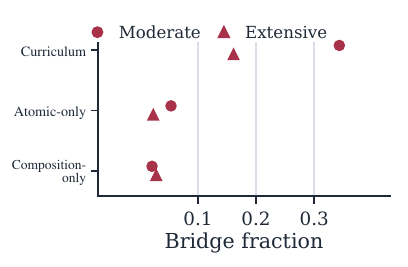}
    \caption{Bridge fraction.}
    \label{fig:mechanism_stage_bridge}
\end{subfigure}\hfill
\begin{subfigure}[t]{0.31\linewidth}
    \centering
    \includegraphics[width=0.98\linewidth]{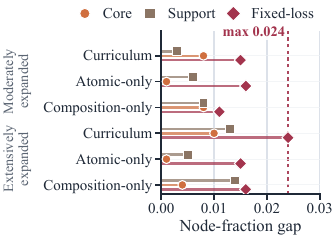}
    \caption{Extraction gaps.}
    \label{fig:mechanism_stage_fraction}
\end{subfigure}
\caption{Compositional gains align with core-circuit bridge fraction. \textbf{A}: curriculum-initialized RL gains concentrate on compositional tasks. \textbf{B}: bridge fraction is the share of compositional-task core-circuit nodes outside matched atomic-task circuits. \textbf{C}: size differences remain small and preserve RL-condition orderings across repeated core extraction and across the two query/key representations of the extended support and fixed-loss pruning graphs (Tables~\ref{tab:mechanism_stage}, \ref{tab:tau_sweep_summary}); expanded-grid endpoint accuracies are reported in Table~\ref{tab:expanded_rl} and Figure~\ref{fig:expanded_rl_endpoints}.}
\label{fig:expanded_rl_behavior}
\end{figure}

\paragraph{Behavioral gains.}
RL initialized from the curriculum-trained supervised checkpoint improves both expanded-distribution families: averaged over five sparsity levels on Suite~1, compositional accuracy rises by $+24.64$ points ($43.83\rightarrow68.47$) in the moderately expanded family and by $+31.27$ points ($46.73\rightarrow78.00$) in the extensively expanded family, while overall accuracy rises from $58.19$ to $73.47$ in the moderately expanded family and to $76.95$ in the extensively expanded family (Figure~\ref{fig:expanded_rl_behavior}; Table~\ref{tab:expanded_rl}).

\paragraph{Bridge fraction.}
The same curriculum-initialized conditions have the largest core-circuit bridge fraction. Atomic-only and composition-only initialization conditions remain weaker behaviorally (their RL Suite~1 mean accuracies are at most $33.37$; Table~\ref{tab:expanded_rl}), while bridge fraction is $0.343$ in the moderately expanded curriculum cell and $0.161$ in the extensively expanded curriculum cell, versus $\le 0.054$ in the other RL conditions (Figure~\ref{fig:expanded_rl_behavior}; Table~\ref{tab:mechanism_stage}). Behavioral recovery on compositional tasks is therefore accompanied by additional core-circuit nodes outside the matched atomic-task circuits---an observed association between behavior and core-circuit structure. Each cell averages over five sparsity levels and three seeds, but the association rests on only the two curriculum cells, which are highest on both compositional accuracy gain and bridge fraction. We therefore read it as a directional pattern across conditions rather than a measured effect size.

\paragraph{Direct bridge intervention.}
We ablate compositional-core nodes outside the matched atomic-core union on dense and $75\%$-sparse SFT and RL checkpoints (training seed 42; five compositional tasks, 200 examples per task). These measurements use teacher-forced tactic-output exact match (EM). At the $75\%$-sparse RL checkpoint, compositional EM falls from $0.975$--$1.000$ to $0.295$--$0.940$, while matched atomic EM remains $0.995$--$1.000$. We also compare distinct bridge and non-bridge subsets matched within node family, layer, and learned selection-score bins. The matched subsets have task-dependent relative selectivity; their results and subset sizes are reported separately from full-set removal (Appendices~\ref{app:rebuttal-bridge} and~\ref{app:bridge-confirmation}).

\begin{takeaway}
\textbf{Takeaway.} The largest exact-match accuracy gains from RL on compositional proofs are accompanied by the largest core-circuit bridge fractions. Full bridge-node ablation in the sparse RL model reduces compositional accuracy while largely preserving accuracy on the corresponding atomic tasks.
\end{takeaway}

Appendix~\ref{app:boundaries} gives boundary and ablation controls that delimit scope and separate atomic-source effects from shared format cues.

\FloatBarrier
\section{Conclusion}

Our comparisons expose different kinds of stability in the same circuit-extraction setting. Across dense and weight-sparse checkpoints, structural overlap is low, while the much higher attention-head overlap under deterministic selection falls sharply when selection among equally scored heads is randomized. Normalized node-fraction gaps are small under the respective extraction comparisons, and each graph preserves its ordering of RL initialization conditions by normalized size. For fixed-loss pruning, Q/K ordering agreement persists through the budget sweep. Finally, the largest exact-match accuracy gains from RL on compositional proofs are accompanied by the largest core-circuit bridge fractions. Full bridge-node ablation in the sparse RL model reduces compositional accuracy while largely preserving accuracy on the corresponding atomic tasks.

Together, these findings isolate a methodological source of ambiguity in circuit extraction. Reproducible graph-size orderings do not by themselves establish that the same components are recovered. The empirical claim supported by a circuit extraction therefore depends on what is extracted and how circuits are compared; preserving behavior alone does not determine that interpretation.

\paragraph{Reporting practice.}
These results suggest a reporting practice for circuit-extraction studies: state the reported graph, the extraction rule and pruning threshold, the scoring and tie-breaking rules, the random baseline, the query/key node representation, the comparison level supporting the claim, and any boundary or negative-control checks that bound the interpretation.

\paragraph{Scope.}
The synthetic Lean family provides fixed proof-state rules and atomic and compositional structures known by construction for auditing circuit extraction in a controlled setting. Extending the same comparisons to broader benchmarks, deeper or larger models, and other architectures is the natural next step; Appendix~\ref{app:limitations_impact} gives the full discussion.

\label{sec:main-text-end}

\section*{Acknowledgements and Disclosures}
The authors received no specific funding for this work. The authors declare no competing interests relevant to this work.

\FloatBarrier
\bibliography{references}
\bibliographystyle{plainnat}

\clearpage
\appendix

\section{Term and Abbreviation Reference}
\label{app:terms}

\begin{table}[H]
\caption{Definitions of paper-specific analysis terms.}
\label{tab:term_abbrev_reference}
\centering
\scriptsize
\setlength{\tabcolsep}{4pt}
\renewcommand{\arraystretch}{1.08}
\paperTableSetup
\begin{tabularx}{\linewidth}{>{\raggedright\arraybackslash}p{0.26\linewidth} X}
\toprule
\paperTableHeader Term & Definition \\
\midrule
Suite~1 / Suite~2 & Two independently sampled evaluation suites generated from the same proof-state generator and task distribution. \\
In-family evaluation & Evaluation on examples sampled from the same proof-state generator and task distribution as the corresponding training/evaluation family, excluding the held-out surface and depth shifts. \\
Core circuit & Compact extracted subgraph whose retained components preserve the target tactic prediction when other candidate components are ablated. \\
Extended support graph & Broader extracted graph that retains additional read, write, and routing components around the same prediction-preserving behavior, beyond the compact core circuit. \\
Fixed-loss pruning graph & The smallest evaluated top-$k$ circuit satisfying the post-ablation loss budget $\tau$, selected from a learned pruning ranking. Appendix~\ref{app:extraction} gives the fallback rules when no evaluated circuit meets the budget. \\
Coupled Q/K & Representation in which query-side and key-side attention support for a head are merged into one \texttt{attn\_qk} node type. \\
Factorized Q/K & Representation in which query-side and key-side attention support are represented separately as \texttt{attn\_q} and \texttt{attn\_k} node types. \\
Structural edge-bundle overlap & Jaccard@10 overlap between groups of support connections defined by layer, source and target component type, and membership in the atomic and compositional circuits; abbreviated as structural overlap. \\
Individual weight-edge overlap & Jaccard@10 overlap between the highest-ranked individual weight connections under path-integrated weight attribution (Appendix~\ref{app:weight_edge_audit}). \\
Routing edge & Selected support edge in the routing relation. For head-level comparisons, routing edges are grouped by the attention head they pass through. \\
Routing-head set & The top-$k$ attention heads ranked by their routing-connection counts. Each head is identified by its layer and head index; equal counts are resolved by the specified selection rule. \\
Random top-$k$ baseline & Uniform sampling with the observed set sizes from the split-level union of observed top-$k$ sets for the same metric. \\
Node fraction & Selected-node count divided by the full node universe for the given graph object and query/key representation. \\
Bridge fraction & Fraction of compositional-task core-circuit nodes that are outside the union of the corresponding atomic-task circuits. \\
Near-miss split & Task--object entries matched by generation seed and selected from non-success cases using the tactic-aware score in Appendix~\ref{app:extraction}. \\
\bottomrule
\end{tabularx}
\paperTableReset
\end{table}

\section{Training Recipes and Checkpoint Selection}
\label{app:training}

This section summarizes the optimization and checkpoint-selection protocols shared across model families within each training stage.

\begin{table}[htbp]
\caption{Training and checkpoint-selection recipes for the model families. Normalization: RMSNorm for SFT; LayerNorm for the GRPO models.}
\label{tab:training_recipes}
\centering
\footnotesize
\setlength{\tabcolsep}{4pt}
\renewcommand{\arraystretch}{1.08}
\paperTableSetup
\begin{tabularx}{\linewidth}{>{\raggedright\arraybackslash}p{0.15\linewidth} >{\raggedright\arraybackslash}p{0.20\linewidth} >{\raggedright\arraybackslash}p{0.30\linewidth} >{\raggedright\arraybackslash}X}
\toprule
\paperTableHeader Stage & Scope & Fixed recipe & Selection / evaluation \\
\midrule
Supervised family & Baseline narrow family and the two expanded-distribution SFT sweeps & Shared 8-layer transformer ($d_{\mathrm{model}}=2048$, $n_{\mathrm{head}}=128$, $d_{\mathrm{head}}=16$, $d_{\mathrm{mlp}}=8192$; $\approx 415.7$M params), 5B tokens, batch size 512, 1000-step warmup, constant $3\times 10^{-4}$ learning rate; curriculum schedule simple-only $[0,0.3)$, gradual mix $[0.3,0.6)$, full task set $[0.6,1.0]$ & Eval every 2000 steps with 50 samples per task; save the best checkpoint by mean Suite~1/Suite~2 eval accuracy; final reporting uses that checkpoint on two independent 100-sample test suites \\
\addlinespace[4pt]
Reinforcement family & Baseline RL sweep and expanded-distribution RL initialization-condition grid & Transformer; GRPO with composition-only prompts, 500 training samples per task, 8 generations per prompt, temperature $1.0$, learning rate $10^{-5}$, $\beta=0.1$, batch size 8 with gradient accumulation 4, and 10 epochs; RL condition labels refer to upstream checkpoint identity (Curriculum, Atomic-only, Composition-only) rather than to prompt mix & Two 50-sample eval suites with early 5-step and later 50-step cadence; the best validation checkpoint is selected by mean Suite~1/Suite~2 eval accuracy; final reporting uses the same two 100-sample test suites \\
\bottomrule
\end{tabularx}
\paperTableReset
\end{table}

GRPO initializes the additional normalization biases to zero. Reported gains compare the SFT and GRPO endpoints.

\begin{table}[htbp]
\caption{Evidence coverage for the main claims and scope checks.}
\label{tab:evidence_scope}
\centering
\footnotesize
\setlength{\tabcolsep}{4pt}
\renewcommand{\arraystretch}{1.06}
\paperTableSetup
\begin{tabularx}{\linewidth}{>{\raggedright\arraybackslash}p{0.29\linewidth} >{\raggedright\arraybackslash}p{0.24\linewidth} >{\raggedright\arraybackslash}X}
\toprule
\paperTableHeader Evidence block & Coverage & Purpose \\
\midrule
Narrow-family behavior (baseline SFT + RL) & 5 sparsities $\times$ 3 seeds & Baseline sparsity and training-stage trends (SFT vs.\ RL) under the baseline narrow generator family \\
\addlinespace[2pt]
Narrow-family graph-level extracted-graph analysis & dense and $75\%$ weight-sparse comparison points, plus supplementary structural-overlap comparisons across all weight-sparse comparison points and cross-seed pairs & Main structural edge-bundle and routing-head overlap comparisons, supplementary checks, and paired dense/weight-sparse bundle comparisons \\
\addlinespace[2pt]
Expanded-distribution supervised behavior & 2 families $\times$ 5 sparsities $\times$ 3 seeds & Supervised effects of training-distribution expansion \\
\addlinespace[2pt]
Expanded-distribution RL behavior & 2 families $\times$ 3 RL initialization conditions $\times$ 5 sparsities $\times$ 3 seeds & RL endpoint performance and accuracy gains where matched SFT baselines are available \\
\addlinespace[2pt]
Expanded-distribution RL extracted-graph analysis & 2 families $\times$ 3 RL initialization conditions $\times$ 5 sparsities $\times$ 3 seeds; repeated core extraction and two Q/K representations for the support and pruning graphs & Condition-level bridge comparisons and graph-size ordering across extraction choices \\
\addlinespace[2pt]
Boundary conditions & 4 regimes $\times$ 2 sparsities & Scope calibration under format shift, distractor shift, and deeper OOD compositions \\
\addlinespace[2pt]
Ranking and null comparisons & 16 task--object pairs per split at the dense--$75\%$-sparse RL comparison & Tie sensitivity and per-side score-weighted structural-bundle baselines \\
\addlinespace[2pt]
Repeated extraction & 2 seed-42 RL checkpoints $\times$ 2 splits; 18 entries per cell & Repeat-extraction overlap, Jaccard and RBO nulls \\
\addlinespace[2pt]
SFT and depth comparisons & 8- and 4-layer SFT; 2 fixed prompt sets, 18 entries per setting/set & Structural-bundle top-10 overlap with size-matched nulls \\
\addlinespace[2pt]
SFT Q/K checks & 12 checkpoints; support and fixed-loss graphs, 5 composition tasks & Normalized node-fraction differences across Q/K representations \\
\addlinespace[2pt]
Bridge interventions & 4 seed-42 checkpoints $\times$ 5 composition tasks; 200 examples/task & Full-set effects and 3 paired subset draws per matching protocol \\
\addlinespace[2pt]
Single-head interventions & 2 seed-42 RL checkpoints $\times$ 9 tasks; 1,024 heads & Mean ablation and counterfactual activation patching \\
\addlinespace[2pt]
Individual weight-edge attribution & 2 seed-42 RL checkpoints $\times$ 9 tasks; 64 pairs per cell & Path-integrated weight attribution, overlap nulls, and ablation checks \\
\bottomrule
\end{tabularx}
\paperTableReset
\end{table}

\FloatBarrier

\section{Circuit Extraction Protocols and Representative-Case Selection}
\label{app:extraction}

Table~\ref{tab:extraction_recipes} summarizes the extraction configurations used for the three object families. The core circuit and extended support graph are both learned-mask extractions that save the mask with the best exact-match target-tactic accuracy after ablation, with the masked-objective value used as a tiebreaker. The loss-constrained pruning subgraph follows the exact-pruning pipeline instead: after optimizing a continuous top-$k$ mask objective, the smallest tested top-$k$ prefix meeting the target-loss budget is selected when available; target-loss satisfaction rates are reported separately. If the extraction search finds no circuit meeting the budget, it retains components with positive learned mask scores and re-evaluates the resulting circuit. In the fixed-ranking loss-budget sweep, an unmet budget instead selects the evaluated candidate with the lowest loss; Table~\ref{tab:tau_sweep_operating_point} reports the budget-satisfaction rates separately. This is not a global minimum over arbitrary subgraphs. This distinction matters for interpretation: the loss-constrained pruning subgraph has a different extraction objective and a different operating constraint from the compact core circuit.

\begin{table}[htbp]
\caption{Circuit-extraction recipes for the three object families. Core extraction uses 200 epochs for the baseline graph analysis and 150 for the expanded-distribution, repeated-extraction, and SFT/depth analyses. Reference activations use a pool generated with 200 candidates per task across all nine tasks, retaining examples that fit the context window. The core circuit uses the same component representation in both runs; the support and pruning graphs compare coupled and factorized Q/K.}
\label{tab:extraction_recipes}
\centering
\footnotesize
\setlength{\tabcolsep}{4pt}
\renewcommand{\arraystretch}{1.07}
\paperTableSetup
\begin{tabularx}{\linewidth}{>{\raggedright\arraybackslash}p{0.17\linewidth} >{\raggedright\arraybackslash}p{0.19\linewidth} >{\raggedright\arraybackslash}p{0.30\linewidth} >{\raggedright\arraybackslash}X}
\toprule
\paperTableHeader Object family & Query/key setup / extractor & Optimization recipe & Discrete object selection \\
\midrule
Core circuit & Attention-output channel and MLP-neuron masks; repeated extraction & Learned hard-concrete mask, $\lambda_{\ell_0}=0.05$, 200 or 150 epochs (see caption), threshold $0.5$, tactic-token supervision, 100 optimization samples per task, 200 reference candidates per task & Save the mask with best exact-match target-tactic accuracy after ablation, with CE+$\lambda_{\ell_0}$ tiebreak; binarize at threshold $0.5$ \\
\addlinespace[3pt]
Extended support graph & Coupled-Q/K support extractor; factorized-Q/K support extractor & Learned read/q/k/v/write mask, $\lambda_{\ell_0}=0.05$, 150 epochs, threshold $0.5$, tactic-token supervision, 100 optimization samples per task, 200 reference candidates per task & Save the mask with best exact-match target-tactic accuracy after ablation, with CE+$\lambda_{\ell_0}$ tiebreak; binarize at threshold $0.5$ \\
\addlinespace[3pt]
Fixed-loss pruning graph & Coupled-Q/K exact-pruning sweep; factorized-Q/K exact-pruning sweep & Exact-pruning objective with $k$-penalty coefficient $3\times 10^{-5}$, 150 epochs, learning rate $3\times 10^{-3}$, weight decay $10^{-3}$, grad clip $1.0$, 100 optimization samples per task, 200 reference candidates per task & After optimization, choose the smallest evaluated top-$k$ circuit meeting the loss budget when available ($\tau=0.15$ for baseline graph extraction; $\tau=0.12$ for expanded-distribution and SFT Q/K comparisons) \\
\bottomrule
\end{tabularx}
\paperTableReset
\end{table}

For the expanded-distribution Q/K comparisons, reference means average batch means. Support fitting uses reference batches of 16 for both representations; pruning fitting uses 16 for coupled Q/K and 32 for factorized Q/K. The fixed-ranking loss-budget sweep uses reference batches of 32 for both.

The representative \textsc{AND-OR} success schematic in Figures~\ref{fig:local_mechanistic_story}, \ref{fig:object_formalization}, and \ref{fig:node_level_routing} follows a fixed representative-case selection protocol. For that task, the protocol evaluates a pre-specified multi-checkpoint pool and summarizes solved cases by target-tactic exact-match score after ablation, membership-category mass from the overlap analysis, highest-mass layer, graph size, and sample length across the core circuit, extended support graph, and loss-constrained pruning subgraph. It then selects the solved case closest to the pool mean after z-scoring these summary features. This protocol chooses illustrative schematics; aggregate evidence comes from the tables.

\paragraph{Near-miss split scoring.}
The structural edge-bundle comparison candidate pool is the pre-generated evaluation pool used for dense/weight-sparse representative-case selection: entries are identified by generation seed. After matching by task, extracted object, and generation seed, and retaining entries with scored bundles available at both endpoints, the near-miss split keeps the top quartile of non-success matched entries by a tactic-aware near-miss score, $0.5$ times mean tactic-line token accuracy plus $0.3$ times first-tactic exact-match rate plus $0.2$ times output-token accuracy.

The intervention and stage/depth comparisons use fixed tokenized inputs shared across endpoints.

Table~\ref{tab:main_claim_map} in the main text gives the compact quantitative summary. The tables that follow provide the underlying evidence blocks, extraction recipes, and fuller quantitative details for each finding family.

\section{Benchmark Tasks and Evaluation Splits}
\label{app:benchmark}

\begin{table}[htbp]
\caption{Atomic and compositional benchmark task families.}
\label{tab:benchmark_tasks}
\centering
\footnotesize
\setlength{\tabcolsep}{4pt}
\renewcommand{\arraystretch}{1.06}
\paperTableSetup
\begin{tabularx}{\linewidth}{>{\raggedright\arraybackslash}p{0.17\linewidth} >{\raggedright\arraybackslash}p{0.18\linewidth} C}
\toprule
\paperTableHeader Family & Task & Goal pattern \\
\midrule
Atomic & \textsc{AND} & $p \land q$ \\
Atomic & \textsc{OR-L} / \textsc{OR-R} & $p \lor q$ \\
Atomic & \textsc{ID} & $p$ \\
Composite & \textsc{AND-OR} & $(p \land q) \lor r$ \\
Composite & \textsc{OR-AND} & $(p \lor q) \land r$ \\
Composite & \textsc{AND-OR-AND} & $(p \land q) \lor (r \land s)$ \\
Composite & \textsc{Nested-AND} & $(p \land q) \land r$ \\
Composite & \textsc{Nested-OR} & $(p \lor q) \lor r$ \\
\bottomrule
\end{tabularx}
\paperTableReset
\end{table}

Table~\ref{tab:benchmark_task_examples} makes the task format concrete. Displayed names are simplified for readability; generated samples randomize proposition names, hypothesis names, order, and distractor premises.

\begin{table}[htbp]
\caption{Canonicalized task examples. Each row shows the proof state given to the model and the target first tactic string; $h_d:d$ is an unused distractor premise.}
\label{tab:benchmark_task_examples}
\centering
\scriptsize
\setlength{\tabcolsep}{3.5pt}
\renewcommand{\arraystretch}{1.08}
\paperTableSetup
\begin{tabularx}{\linewidth}{>{\raggedright\arraybackslash}p{0.12\linewidth} >{\raggedright\arraybackslash}p{0.14\linewidth} >{\raggedright\arraybackslash}p{0.42\linewidth} >{\raggedright\arraybackslash}p{0.24\linewidth}}
\toprule
\paperTableHeader Family & Task & Example proof state & Target tactic \\
\midrule
Atomic & \textsc{AND} &
\begin{tabular}[t]{@{}l@{}}
$p,q,d:\mathrm{Prop}$\\
$h_p:p,\ h_q:q,\ h_d:d$\\
$\vdash p \land q$
\end{tabular} &
\texttt{exact And.intro h\_p h\_q} \\
Atomic & \textsc{OR-L} &
\begin{tabular}[t]{@{}l@{}}
$p,q,d:\mathrm{Prop}$\\
$h_p:p,\ h_d:d$\\
$\vdash p \lor q$
\end{tabular} &
\texttt{exact Or.inl h\_p} \\
Atomic & \textsc{ID} &
\begin{tabular}[t]{@{}l@{}}
$p,d:\mathrm{Prop}$\\
$h_p:p,\ h_d:d$\\
$\vdash p$
\end{tabular} &
\texttt{exact h\_p} \\
Composite & \textsc{AND-OR} &
\begin{tabular}[t]{@{}l@{}}
$p,q,r,d:\mathrm{Prop}$\\
$h_p:p,\ h_q:q,\ h_d:d$\\
$\vdash (p \land q) \lor r$
\end{tabular} &
\texttt{exact Or.inl (And.intro h\_p h\_q)} \\
Composite & \textsc{OR-AND} &
\begin{tabular}[t]{@{}l@{}}
$p,q,r,d:\mathrm{Prop}$\\
$h_p:p,\ h_r:r,\ h_d:d$\\
$\vdash (p \lor q) \land r$
\end{tabular} &
\texttt{exact And.intro (Or.inl h\_p) h\_r} \\
Composite & \textsc{Nested-OR} &
\begin{tabular}[t]{@{}l@{}}
$p,q,r,d:\mathrm{Prop}$\\
$h_p:p,\ h_d:d$\\
$\vdash (p \lor q) \lor r$
\end{tabular} &
\texttt{exact Or.inl (Or.inl h\_p)} \\
\bottomrule
\end{tabularx}
\paperTableReset
\end{table}

Suite~1 and Suite~2 are two independently generated test suites. Reporting both reduces the risk of overstating a single draw from the dynamic generator.

\FloatBarrier

\section{Supplementary Behavioral Tables}
\label{app:behavior_tables}

\begin{table}[htbp]
\caption{Baseline-distribution family across sparsity. Means are computed over three seeds.}
\label{tab:baseline_behavior}
\centering
\footnotesize
\setlength{\tabcolsep}{4pt}
\renewcommand{\arraystretch}{1.06}
\paperTableSetup
\begin{tabularx}{\linewidth}{>{\raggedright\arraybackslash}p{0.18\linewidth} C C C C}
\toprule
\paperTableHeader Sparsity & SFT Suite~1 & SFT Suite~2 & RL Suite~1 & RL Suite~2 \\
\midrule
Dense & 59.22 & 59.41 & 93.93 & 93.37 \\
$25\%$-sparse & 67.19 & 66.04 & 93.89 & 93.52 \\
$50\%$-sparse & 66.56 & 68.00 & 98.89 & 99.15 \\
$75\%$-sparse & 68.48 & 71.30 & 96.44 & 96.67 \\
$90\%$-sparse & 74.41 & 74.00 & 98.37 & 98.26 \\
\bottomrule
\end{tabularx}
\paperTableReset
\end{table}

\begin{table}[htbp]
\caption{Expanded-distribution supervised families across sparsity. Means are computed over three seeds.}
\label{tab:expanded_supervised}
\centering
\footnotesize
\setlength{\tabcolsep}{4pt}
\renewcommand{\arraystretch}{1.06}
\paperTableSetup
\begin{tabularx}{\linewidth}{>{\raggedright\arraybackslash}p{0.18\linewidth} C C C C}
\toprule
\paperTableHeader Sparsity & Moderate Suite~1 & Moderate Suite~2 & Extensive Suite~1 & Extensive Suite~2 \\
\midrule
Dense & 59.26 & 58.11 & 57.04 & 55.85 \\
$25\%$-sparse & 55.96 & 55.70 & 63.41 & 63.93 \\
$50\%$-sparse & 54.33 & 54.00 & 52.63 & 54.37 \\
$75\%$-sparse & 60.56 & 60.52 & 56.41 & 56.96 \\
$90\%$-sparse & 60.85 & 60.56 & 61.48 & 60.11 \\
\bottomrule
\end{tabularx}
\paperTableReset
\end{table}

\begin{table}[htbp]
\caption{Expanded-distribution RL condition summary across all sparsity and seed settings, reporting Suite~1 and compositional-task accuracy means. Means are aggregated over 5 sparsity levels $\times$ 3 seeds within each family-condition cell. Condition labels denote the upstream supervised checkpoint used to initialize the same composition-only GRPO refinement.}
\label{tab:expanded_rl}
\centering
\footnotesize
\setlength{\tabcolsep}{4pt}
\renewcommand{\arraystretch}{1.06}
\paperTableSetup
\begin{tabularx}{\linewidth}{
>{\raggedright\arraybackslash}p{0.22\linewidth}
>{\raggedright\arraybackslash}p{0.17\linewidth}
C C C C C C}
\toprule
\paperTableHeader Family & Condition & SFT Suite~1 & RL Suite~1 & $\Delta$ Suite~1 & SFT comp. & RL comp. & $\Delta$ comp. \\
\midrule
Moderately expanded & Curriculum & 58.19 & 73.47 & 15.28 & 43.83 & 68.47 & 24.64 \\
Moderately expanded & Atomic-only & --- & 29.94 & --- & --- & 4.27 & $\le 4.27$ \\
Moderately expanded & Composition-only & --- & 29.74 & --- & 36.36 & 50.56 & 14.20 \\
Extensively expanded & Curriculum & 58.19 & 76.95 & 18.76 & 46.73 & 78.00 & 31.27 \\
Extensively expanded & Atomic-only & --- & 28.29 & --- & --- & 5.59 & $\le 5.59$ \\
Extensively expanded & Composition-only & --- & 33.37 & --- & 39.69 & 57.80 & 18.11 \\
\bottomrule
\end{tabularx}
\vspace{2pt}
\parbox{\linewidth}{\scriptsize Suite~1 means cover all nine task types. The available SFT evaluations for Atomic-only and Composition-only initialization cover only the four atomic tasks and the five compositional tasks, respectively. Dashes indicate unavailable SFT accuracies or matched gains. Atomic-only compositional gains are bounded above by the RL accuracies because SFT accuracy is nonnegative. Atomic-only SFT atomic-task means are $55.42/50.12$ for the moderate/extensive distributions.}
\paperTableReset
\end{table}

\begin{figure}[htbp]
\centering
\includegraphics[width=0.48\linewidth]{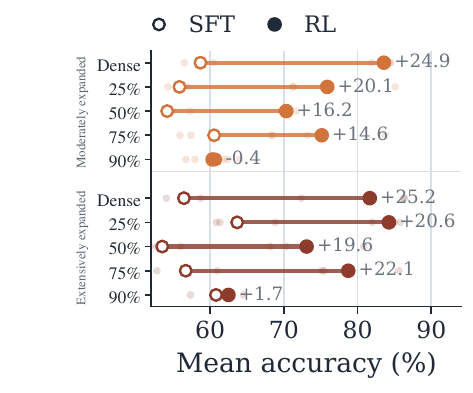}
\caption{Expanded-grid RL endpoint accuracies for curriculum-initialized models in the moderately and extensively expanded distributions. Bars compare the upstream supervised checkpoint with the RL-refined endpoint across sparsity levels, using mean accuracy across Suites~1 and~2; annotations report the corresponding RL--SFT accuracy changes. Table~\ref{tab:expanded_rl} separately reports Suite~1 and compositional-task means by initialization condition.}
\label{fig:expanded_rl_endpoints}
\end{figure}

\FloatBarrier

\section{Supplementary Extracted-Graph Tables}
\label{app:mechanism_tables}

This section provides the numerical counterparts to the main extracted-graph findings. Table~\ref{tab:edge_overlap} records the exact dense-versus-weight-sparse overlap values, Table~\ref{tab:edge_overlap_random} adds a random top-$k$ baseline using the observed candidate set, Table~\ref{tab:mechanism_stage} records the condition-by-condition object statistics underlying Figure~\ref{fig:expanded_rl_behavior}, and Tables~\ref{tab:mechanism_stage_ordering}--\ref{tab:mechanism_stage_bootstrap} summarize how much the RL-condition ordering itself is preserved across repeated core extractions and across coupled and factorized Q/K representations of the extended support and fixed-loss pruning graphs.

\begin{table}[htbp]
\caption{Structural edge-bundle and routing-head overlap between the dense RL checkpoint and the RL checkpoint at $75\%$ weight sparsity. Routing-head sets use deterministic selection within tied score groups. Confidence intervals are 95\% bootstrap intervals over the 16 matched task--object entries per split. Semantic overlap is reported for completeness; in these comparisons it is numerically indistinguishable from structural overlap.}
\label{tab:edge_overlap}
\centering
\footnotesize
\setlength{\tabcolsep}{4pt}
\renewcommand{\arraystretch}{1.06}
\paperTableSetup
\begin{tabularx}{\linewidth}{>{\raggedright\arraybackslash}p{0.17\linewidth} C C C}
\toprule
\paperTableHeader Split & Structural Jaccard@10 & Semantic Jaccard@10 & Routing-head-set Jaccard@10 \\
\midrule
Success & $0.163\ [0.135, 0.195]$ & $0.163\ [0.135, 0.194]$ & $0.666\ [0.586, 0.748]$ \\
Near-miss & $0.142\ [0.111, 0.177]$ & $0.142\ [0.109, 0.177]$ & $0.554\ [0.479, 0.631]$ \\
\bottomrule
\end{tabularx}
\paperTableReset
\end{table}

\begin{table}[htbp]
\caption{Uniform random top-$k$ baseline for the dense-vs-$75\%$-weight-sparse structural and routing-head overlap comparison. For each split and metric, the random baseline draws top-$k$ sets with the observed set sizes from the split-level union of the observed top-$k$ sets for that metric. We use 50,000 Monte Carlo draws; $p=(1+\#\{\bar J_{\rm null}\geq\bar J_{\rm obs}\})/50{,}001$, where $\bar J$ is the split-level mean overlap. Routing observations use deterministic selection within tied score groups. Appendix~\ref{app:rebuttal-ranking} reports a separate 2,000-draw comparison using uniform and score-weighted structural baselines, together with sensitivity to random tie-breaking.}
\label{tab:edge_overlap_random}
\centering
\footnotesize
\setlength{\tabcolsep}{4pt}
\renewcommand{\arraystretch}{1.04}
\paperTableSetup
\begin{tabular}{>{\raggedright\arraybackslash}p{0.10\linewidth} >{\raggedright\arraybackslash}p{0.13\linewidth} >{\centering\arraybackslash}p{0.080\linewidth} >{\centering\arraybackslash}p{0.10\linewidth} >{\centering\arraybackslash}p{0.20\linewidth} >{\centering\arraybackslash}p{0.10\linewidth} >{\centering\arraybackslash}p{0.07\linewidth}}
\toprule
\paperTableHeader Split & Metric & Observed & Random mean & Random 95\% interval & $p$ & Cand.\ set \\
\midrule
Success & Structural & 0.163 & 0.126 & $[0.087, 0.168]$ & 0.040 & 37 \\
Success & Semantic & 0.163 & 0.126 & $[0.088, 0.168]$ & 0.039 & 37 \\
Success & Routing-head sets & 0.666 & 0.199 & $[0.156, 0.245]$ & $<0.001$ & 31 \\
Near-miss & Structural & 0.142 & 0.117 & $[0.080, 0.158]$ & 0.105 & 39 \\
Near-miss & Semantic & 0.142 & 0.117 & $[0.080, 0.158]$ & 0.108 & 39 \\
Near-miss & Routing-head sets & 0.554 & 0.185 & $[0.143, 0.230]$ & $<0.001$ & 33 \\
\bottomrule
\end{tabular}
\paperTableReset
\end{table}

\begin{table}[htbp]
\caption{Extracted-graph statistics by RL initialization condition across expanded-distribution families. Means are aggregated over 5 sparsity levels $\times$ 3 seeds within each family-condition cell; condition labels denote the upstream supervised checkpoint used to initialize the same composition-only GRPO refinement. Core fraction and core bridge fraction use the first core extraction. The gap columns report mean absolute node-fraction differences: between repeated extractions for the core circuit, and between coupled and factorized Q/K representations for the extended support and fixed-loss pruning graphs. Fractions are computed as selected-node counts divided by each object's node universe, so these columns are within-object summaries instead of common-scale effect sizes across objects.}
\label{tab:mechanism_stage}
\centering
\scriptsize
\setlength{\tabcolsep}{1.8pt}
\renewcommand{\arraystretch}{1.08}
\paperTableSetup
\begin{tabularx}{\linewidth}{@{}
>{\raggedright\arraybackslash}p{0.14\linewidth}
>{\raggedright\arraybackslash}p{0.15\linewidth}
>{\centering\arraybackslash}p{0.10\linewidth}
>{\centering\arraybackslash}p{0.09\linewidth}
>{\centering\arraybackslash}p{0.10\linewidth}
>{\centering\arraybackslash}p{0.11\linewidth}
C@{}}
\toprule
\paperTableHeader Family & Condition & \shortstack[c]{Core\\fraction} & \shortstack[c]{Core bridge\\fraction} & \shortstack[c]{Core repeat\\gap} & \shortstack[c]{Support\\gap} & \shortstack[c]{Loss-constr.\\gap} \\
\midrule
Moderately expanded & Curriculum & 0.514 & 0.343 & 0.008 & 0.003 & 0.015 \\
Moderately expanded & Atomic-only & 0.773 & 0.054 & 0.001 & 0.006 & 0.016 \\
Moderately expanded & Composition-only & 0.721 & 0.021 & 0.008 & 0.008 & 0.011 \\
Extensively expanded & Curriculum & 0.674 & 0.161 & 0.010 & 0.013 & 0.024 \\
Extensively expanded & Atomic-only & 0.776 & 0.023 & 0.001 & 0.005 & 0.015 \\
Extensively expanded & Composition-only & 0.710 & 0.028 & 0.004 & 0.014 & 0.016 \\
\bottomrule
\end{tabularx}
\paperTableReset
\end{table}

\begin{table}[htbp]
\caption{RL initialization-condition ordering agreement across expanded-distribution families: repeated extraction for the core circuit, and coupled versus factorized Q/K representations for the extended support and fixed-loss pruning graphs.}
\label{tab:mechanism_stage_ordering}
\centering
\footnotesize
\setlength{\tabcolsep}{4pt}
\renewcommand{\arraystretch}{1.06}
\paperTableSetup
\begin{tabularx}{\linewidth}{>{\raggedright\arraybackslash}p{0.28\linewidth} C C C}
\toprule
\paperTableHeader Family & Core circuit & Extended support & Loss-constrained \\
\midrule
Moderately expanded & 3/3 & 3/3 & 3/3 \\
Extensively expanded & 3/3 & 3/3 & 3/3 \\
\bottomrule
\end{tabularx}
\paperTableReset
\end{table}

\begin{table}[htbp]
\caption{Bootstrap check for the RL initialization-condition ordering result. Each bootstrap sample resamples the 15 sparsity-seed units within each family-condition cell. The last two columns report the probability of disagreement on the two orderings involving the curriculum-initialized condition, across repeated core extraction or across coupled and factorized Q/K representations of the extended support and fixed-loss pruning graphs.}
\label{tab:mechanism_stage_bootstrap}
\centering
\footnotesize
\setlength{\tabcolsep}{4pt}
\renewcommand{\arraystretch}{1.05}
\paperTableSetup
\begin{tabularx}{\linewidth}{>{\raggedright\arraybackslash}p{0.22\linewidth} >{\raggedright\arraybackslash}p{0.24\linewidth} C C C}
\toprule
\paperTableHeader Family & Object & Observed agreement & \shortstack[c]{Flip prob.\\Curric.\ vs Atomic} & \shortstack[c]{Flip prob.\\Curric.\ vs Comp.} \\
\midrule
Moderately expanded & Core circuit & 3/3 & 0.000 & 0.000 \\
Moderately expanded & Extended support & 3/3 & 0.000 & 0.000 \\
Moderately expanded & Loss-constrained & 3/3 & 0.005 & 0.095 \\
Extensively expanded & Core circuit & 3/3 & 0.000 & 0.099 \\
Extensively expanded & Extended support & 3/3 & 0.000 & 0.027 \\
Extensively expanded & Loss-constrained & 3/3 & 0.003 & 0.232 \\
\bottomrule
\end{tabularx}
\paperTableReset
\end{table}

\begin{table}[htbp]
\caption{Loss-budget sweep for the fixed-loss pruning graph, holding the fitted pruning-score ranking fixed. For each family, loss budget, and curriculum-initialized condition, the coupled-vs-factorized query/key gap is computed by averaging node fractions over seeds and compositional tasks within each sparsity level, taking the coupled-vs-factorized absolute difference at each sparsity, then averaging those five sparsity-level gaps. Pairwise RL-condition ordering agreement between coupled and factorized is reported at five tested loss budgets. The main reporting point used in Section~\ref{sec:setup_dependence} and Tables~\ref{tab:mechanism_stage}--\ref{tab:mechanism_stage_bootstrap} is $\tau=0.12$. Across the tested range, gaps stay small in absolute terms and ordering agreement stays at $3/3$ for both expanded distributions. In both families and both Q/K representations, the descending node-fraction order is Curriculum $>$ Atomic-only $>$ Composition-only at $\tau=0.04$, and Atomic-only $>$ Curriculum $>$ Composition-only at $\tau\in\{0.08,0.12,0.16,0.20\}$.}
\label{tab:tau_sweep_summary}
\centering
\footnotesize
\setlength{\tabcolsep}{4pt}
\renewcommand{\arraystretch}{1.05}
\paperTableSetup
\begin{tabularx}{\linewidth}{>{\raggedright\arraybackslash}p{0.20\linewidth} C C C C C}
\toprule
\paperTableHeader Family / Quantity & $\tau=0.04$ & $\tau=0.08$ & $\tau=0.12$ & $\tau=0.16$ & $\tau=0.20$ \\
\midrule
Mod.\ exp.\ curric.\ gap & 0.050 & 0.030 & 0.015 & 0.017 & 0.013 \\
Ext.\ exp.\ curric.\ gap & 0.055 & 0.020 & 0.024 & 0.013 & 0.013 \\
Mod.\ exp.\ ordering agr.\ & 3/3 & 3/3 & 3/3 & 3/3 & 3/3 \\
Ext.\ exp.\ ordering agr.\ & 3/3 & 3/3 & 3/3 & 3/3 & 3/3 \\
\bottomrule
\end{tabularx}
\paperTableReset
\end{table}

\begin{table}[htbp]
\caption{Pruning-threshold diagnostics for the loss-budget sweep. As the allowed post-ablation loss increases, the selected fixed-loss pruning graph becomes smaller and satisfies the target-loss constraint more often. These aggregate diagnostics are computed from the same loss-budget sweep as Table~\ref{tab:tau_sweep_summary} and serve as a pruning-threshold check, not as an additional finding.}
\label{tab:tau_sweep_operating_point}
\centering
\footnotesize
\setlength{\tabcolsep}{5pt}
\renewcommand{\arraystretch}{1.05}
\paperTableSetup
\begin{tabularx}{\linewidth}{C C C C C}
\toprule
\paperTableHeader Loss budget $\tau$ & Target-loss satisfaction rate & Mean node fraction & Median node fraction & Mean selected size $k$ \\
\midrule
0.04 & 0.299 & 0.362 & 0.303 & 62{,}173 \\
0.08 & 0.627 & 0.276 & 0.236 & 47{,}403 \\
0.12 & 0.711 & 0.246 & 0.213 & 42{,}232 \\
0.16 & 0.738 & 0.240 & 0.211 & 41{,}195 \\
0.20 & 0.763 & 0.236 & 0.209 & 40{,}566 \\
\bottomrule
\end{tabularx}
\paperTableReset
\end{table}

\FloatBarrier

\section{Supplementary Edge-Overlap Comparisons Across Sparsity and Seed}
\label{app:exact_edge_supplement}

\begin{figure}[p]
\centering
\begin{subfigure}[t]{0.78\linewidth}
    \centering
    \includegraphics[width=\linewidth]{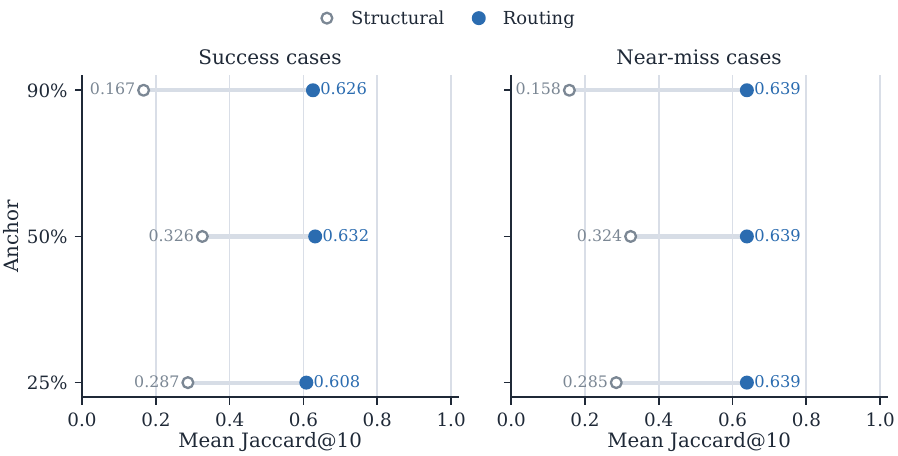}
    \caption{Cross-sparsity dense-vs-weight-sparse comparisons.}
    \label{fig:exact_edge_supplement_cross_sparsity}
\end{subfigure}

\vspace{0.5em}

\begin{subfigure}[t]{0.78\linewidth}
    \centering
    \includegraphics[width=\linewidth]{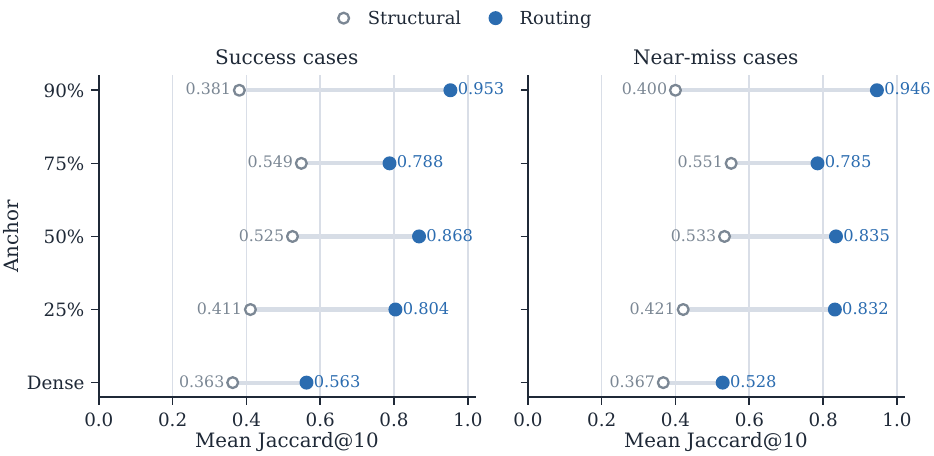}
    \caption{Matched-sparsity cross-seed comparisons.}
    \label{fig:exact_edge_supplement_cross_seed}
\end{subfigure}
\caption{Structural and routing-head overlap beyond the main dense and $75\%$ weight-sparse comparison point. Under deterministic selection, routing-head-set Jaccard@10 remains higher than structural Jaccard@10 across both cross-sparsity and cross-seed comparisons. Semantic overlap exactly matches structural overlap throughout this supplement. Figure~\ref{fig:edge_overlap} compares selection rules at the main dense--sparse pair.}
\label{fig:exact_edge_supplement}
\end{figure}

Table~\ref{tab:exact_edge_supplement} and Figure~\ref{fig:exact_edge_supplement} extend the dense-versus-$75\%$-weight-sparse comparison to other sparsity levels and matched-sparsity checkpoint pairs across seeds. Under deterministic selection, routing-head Jaccard@10 remains higher than structural Jaccard@10 in these comparisons. The table caption defines the comparison units; Appendix~\ref{app:rebuttal-ranking} examines sensitivity to selection within tied score groups.

\begin{table}[htbp]
\caption{Supplementary structural and routing-head overlap comparisons beyond the main dense and $75\%$ weight-sparse comparison point. Cross-sparsity rows compare dense against the other weight-sparse checkpoints within the matched task-object entries; each row summarizes one checkpoint pair ($n=1$). Cross-seed rows compare matched-sparsity pairs across seeds; each row averages three seed pairs ($n=3$). Values are mean Jaccard@10 under deterministic selection; $n$ counts checkpoint pairs rather than task-object entries.}
\label{tab:exact_edge_supplement}
\centering
\footnotesize
\setlength{\tabcolsep}{4pt}
\renewcommand{\arraystretch}{1.04}
\paperTableSetup
\begin{tabularx}{\linewidth}{>{\raggedright\arraybackslash}p{0.34\linewidth} >{\raggedright\arraybackslash}p{0.14\linewidth} C C C C}
\toprule
\paperTableHeader Setting & Split & Structural & Semantic & Routing & $n$ \\
\midrule
Cross-sparsity, $25\%$-sparse & Success & 0.287 & 0.287 & 0.608 & 1 \\
Cross-sparsity, $50\%$-sparse & Success & 0.326 & 0.326 & 0.632 & 1 \\
Cross-sparsity, $90\%$-sparse & Success & 0.167 & 0.167 & 0.626 & 1 \\
Cross-sparsity, $25\%$-sparse & Near-miss & 0.285 & 0.285 & 0.639 & 1 \\
Cross-sparsity, $50\%$-sparse & Near-miss & 0.324 & 0.324 & 0.639 & 1 \\
Cross-sparsity, $90\%$-sparse & Near-miss & 0.158 & 0.158 & 0.639 & 1 \\
Cross-seed, Dense & Success & 0.363 & 0.363 & 0.563 & 3 \\
Cross-seed, $25\%$-sparse & Success & 0.411 & 0.411 & 0.804 & 3 \\
Cross-seed, $50\%$-sparse & Success & 0.525 & 0.525 & 0.868 & 3 \\
Cross-seed, $75\%$-sparse & Success & 0.549 & 0.549 & 0.788 & 3 \\
Cross-seed, $90\%$-sparse & Success & 0.381 & 0.381 & 0.953 & 3 \\
Cross-seed, Dense & Near-miss & 0.367 & 0.367 & 0.528 & 3 \\
Cross-seed, $25\%$-sparse & Near-miss & 0.421 & 0.421 & 0.832 & 3 \\
Cross-seed, $50\%$-sparse & Near-miss & 0.533 & 0.533 & 0.835 & 3 \\
Cross-seed, $75\%$-sparse & Near-miss & 0.551 & 0.551 & 0.785 & 3 \\
Cross-seed, $90\%$-sparse & Near-miss & 0.400 & 0.400 & 0.946 & 3 \\
\bottomrule
\end{tabularx}
\paperTableReset
\end{table}

\FloatBarrier

\makeatletter
\setlength{\@fptop}{0pt}
\setlength{\@fpsep}{10pt}
\setlength{\@fpbot}{0pt plus 1fil}
\makeatother

\FloatBarrier

\section{Boundary Conditions and Negative Controls}
\label{app:boundaries}

\begin{figure}[htbp]
\centering
\begin{subfigure}[t]{0.47\linewidth}
    \centering
    \includegraphics[width=\linewidth]{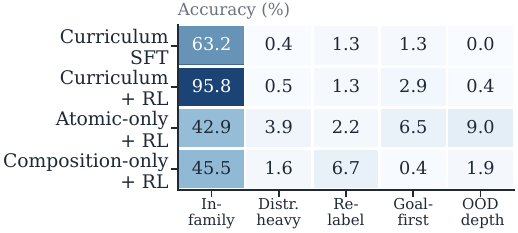}
    \caption{Boundary-condition behavior.}
    \label{fig:boundary_conditions}
\end{subfigure}\hfill
\begin{subfigure}[t]{0.47\linewidth}
    \centering
    \includegraphics[width=\linewidth]{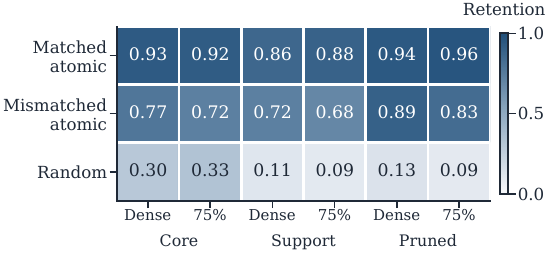}
    \caption{Target predictions retained by atomic-control circuits.}
    \label{fig:negative_controls}
\end{subfigure}
\caption{Boundary tests and controls for the narrow-family regimes at the dense and $75\%$ weight-sparse comparison points. \textbf{A}: the strongest baseline-family behavior does not carry over to the evaluated held-out surface/depth shifts. \textbf{B}: matched atomic unions preserve the most target predictions after ablation, and mismatched atomic unions remain well above random controls. The matched-vs-mismatched gap indicates that the correct atomic source matters; the mismatched-vs-random gap is consistent with shared format/task-family cues contributing.}
\label{fig:boundary_scope}
\end{figure}

These tables provide behavior-level scope and control evidence in raw numeric form. Boundary tests calibrate the scope of the in-family findings reported in Section~\ref{sec:results}: the curriculum-initialized condition (labeled Curriculum + RL in Table~\ref{tab:boundary_conditions}) reaches $95.79\%$ in-family accuracy but only $0.51\%$ on the distractor shift, $1.30\%$ on the relabel shift, $2.92\%$ on the goal-first shift, and $0.42\%$ on the deeper compositional-depth check (OOD depth; Figure~\ref{fig:boundary_scope}, Table~\ref{tab:boundary_conditions}). The extraction findings characterize the in-family proof-state generator and should not be interpreted as evidence of OOD transfer or broader compositional generalization. Within this in-family setting, matched atomic unions retain more target predictions after ablation than mismatched atomic unions, while mismatched unions remain well above random controls (Table~\ref{tab:negative_controls}). The matched-vs-mismatched gap indicates that the correct atomic source matters; the mismatched-vs-random gap is consistent with shared format/task-family cues contributing.

\begin{table}[htbp]
\caption{Boundary-condition summary for narrow-family regimes at the dense and $75\%$ weight-sparse comparison points. Values are regime means in percent. Condition labels denote the upstream checkpoint used before the same RL refinement: ``Atomic-only + RL'' starts from the atomic-only supervised checkpoint, and ``Composition-only + RL'' starts from the composition-only supervised checkpoint. The in-family column is the standard evaluation; Distractor / Relabel / Goal-first are held-out surface or format shifts; OOD depth is deeper compositional extrapolation.}
\label{tab:boundary_conditions}
\centering
\footnotesize
\setlength{\tabcolsep}{4pt}
\renewcommand{\arraystretch}{1.06}
\paperTableSetup
\begin{tabularx}{\linewidth}{>{\raggedright\arraybackslash}p{0.26\linewidth} C C C C C}
\toprule
\paperTableHeader Regime & In-family & Distractor & Relabel & Goal-first & OOD depth \\
\midrule
Curriculum SFT & 63.24 & 0.37 & 1.30 & 1.34 & 0.00 \\
Curriculum + RL & 95.79 & 0.51 & 1.30 & 2.92 & 0.42 \\
Atomic-only + RL & 42.87 & 3.89 & 2.22 & 6.48 & 8.96 \\
Composition-only + RL & 45.46 & 1.62 & 6.67 & 0.42 & 1.88 \\
\bottomrule
\end{tabularx}
\paperTableReset
\end{table}

\begin{table}[htbp]
\caption{Negative-control target-prediction retention after ablation, aggregated over five compositional tasks and three seeds. ``Matched atomic'' keeps the union of the atomic-task circuits that define the compositional task; ``mismatched atomic'' keeps a low-overlap atomic union from the same checkpoint; ``random'' averages size-, layer-, and layer-family-matched random controls.}
\label{tab:negative_controls}
\centering
\scriptsize
\setlength{\tabcolsep}{2.8pt}
\renewcommand{\arraystretch}{1.06}
\paperTableSetup
\begin{tabularx}{\linewidth}{>{\raggedright\arraybackslash}p{0.27\linewidth} C C C C C}
\toprule
\paperTableHeader Object / Checkpoint & Matched atomic & Mismatched atomic & Random & \shortstack[c]{Matched\\$-$mismatch} & \shortstack[c]{Mismatch\\$-$random} \\
\midrule
Core circuit / Dense & 0.926 & 0.774 & 0.298 & 0.152 & 0.476 \\
Core circuit / $75\%$-sparse & 0.920 & 0.722 & 0.332 & 0.198 & 0.390 \\
Extended support graph / Dense & 0.859 & 0.719 & 0.110 & 0.140 & 0.609 \\
Extended support graph / $75\%$-sparse & 0.884 & 0.680 & 0.095 & 0.204 & 0.585 \\
Loss-constrained pruning subgraph / Dense & 0.937 & 0.893 & 0.133 & 0.044 & 0.760 \\
Loss-constrained pruning subgraph / $75\%$-sparse & 0.956 & 0.830 & 0.092 & 0.126 & 0.737 \\
\bottomrule
\end{tabularx}
\paperTableReset
\end{table}

\section{Additional Schematic Examples}
\label{app:schematic}

\begin{figure}[tbp]
\centering
\includegraphics[width=\linewidth]{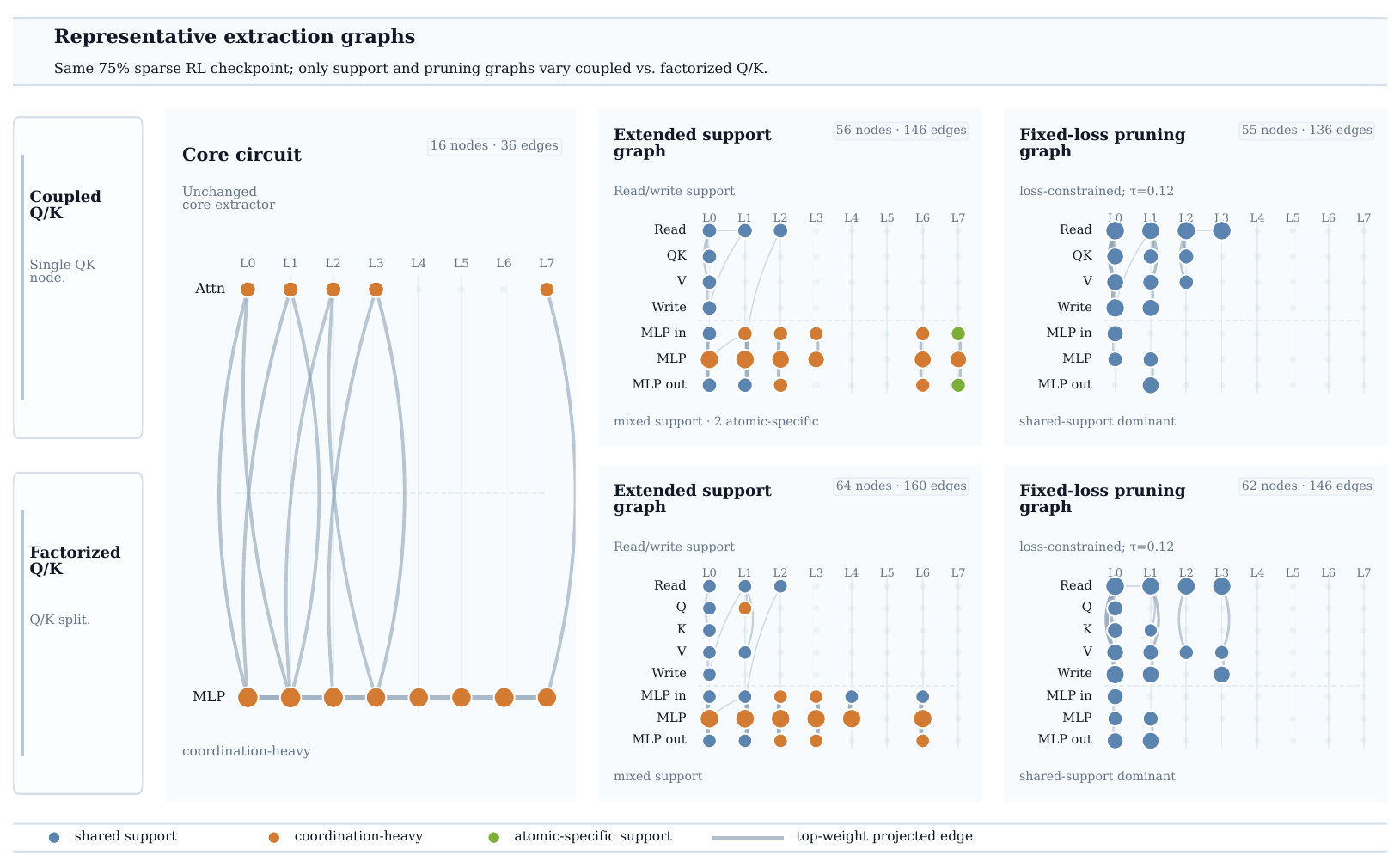}
\caption{Graph-level counterpart to Figure~\ref{fig:object_formalization}. Columns compare the core circuit, extended support graph, and fixed-loss pruning graph; the support and pruning columns compare coupled Q/K and factorized Q/K on the same representative checkpoint, while the core circuit is shown once. Displayed nodes group selected components by layer and component type, and links summarize connections between these groups. Node and edge counts refer to this displayed summary.}
\label{fig:object_formalization_graph_matrix}
\end{figure}

\begin{figure}[tbp]
\centering
\includegraphics[width=\linewidth]{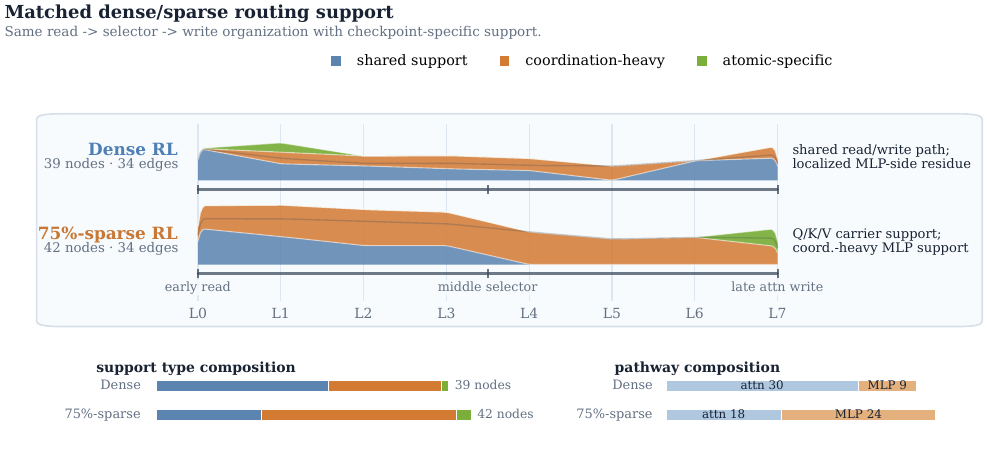}
\caption{Paired dense/weight-sparse routing support for the representative \textsc{AND-OR} success case discussed in Section~\ref{sec:routing_anchor}. Both the dense and $75\%$ weight-sparse RL endpoints preserve the same coarse early-read $\rightarrow$ middle-selector $\rightarrow$ late-attention-write organization, but the supporting components differ. The lower bars summarize support-category and attention-vs-MLP composition; the displayed weight-sparse graph has a larger MLP share than the dense graph. Aggregate overlap comparisons are reported in Figure~\ref{fig:edge_overlap}.}
\label{fig:node_level_routing}
\end{figure}

The next two figures play different supporting roles. Figure~\ref{fig:object_formalization_graph_matrix} shows the graph structure for the same representative checkpoint summarized in Figure~\ref{fig:object_formalization}. Figure~\ref{fig:node_level_routing} provides the paired dense/weight-sparse routing comparison for the representative success case discussed in Section~\ref{sec:routing_anchor}.

\FloatBarrier

\section{Limitations and Broader Impact}
\label{app:limitations_impact}

\paragraph{Benchmark and model scope.}
The synthetic Lean benchmark, eight-layer 415.7M-parameter main model, and four-layer SFT pilot form a deliberately controlled setting: explicit atomic and compositional proof steps, dynamic names and distractors, two independently generated evaluation suites, and matched training recipes are precisely what enable controlled circuit comparison, while the model is large enough for nontrivial sparse and RL behavior yet tractable enough for repeated extraction and seed/sparsity comparisons. The shared proof-state generation rules may also make format-level cues shared across tasks easier to observe than in more heterogeneous families. Validating the same sensitivity to extraction choices at larger scales, in other architectures, and across more heterogeneous benchmarks is the natural next step. The full sparsity sweep in Figure~\ref{fig:supervised_behavior_maps} also provides context for the graph-extraction anchors: under the fixed training recipe used here, accuracy is not monotonic in weight sparsity.

\paragraph{Graph evidence and pruning thresholds.}
The main overlap comparison in the baseline distribution covers $16+16$ matched task--object entries at the pre-specified dense and $75\%$ weight-sparse checkpoints; representative graph visualizations illustrate individual cases rather than estimate their prevalence across extracted graphs. Table~\ref{tab:exact_edge_supplement} and Figure~\ref{fig:exact_edge_supplement} extend the overlap checks across all weight-sparse checkpoints and cross-seed pairs. The loss-constrained pruning subgraph also depends on its pruning threshold, since its discrete object is selected under an explicit loss budget; the main reporting point is $\tau=0.12$, and Table~\ref{tab:tau_sweep_summary} extends the coupled-vs-factorized comparison to four additional loss budgets.

\paragraph{Boundary tests and negative controls.}
Boundary tests in Appendix~\ref{app:boundaries} measure behavior under shift, and circuit-level stability under shift remains a separate validation target. Negative controls show that matched atomic controls preserve more target predictions after ablation than mismatched controls, while mismatched controls remain well above random; this pattern is consistent with shared format/task-family cues contributing. Fixed-checkpoint protocol audits isolate specific analysis choices; cross-checkpoint comparisons also include model differences. Broader behavioral transfer remains unvalidated.

\paragraph{Scope of circuit-style claims.}
Beyond the specific choices audited here, recent field-level assessments argue that fully reverse-engineering the internal computation of large models may be unrealistic, and that interpretability should instead target partial understanding that is validated on downstream tasks \citep{Nanda2025PragmaticVision}. Our results do not resolve this debate, but they are consistent with its premise: even in a controlled setting where the task structure is known by construction, overlap at both the structural and attention-head levels depends on how components are selected and compared. The reporting practice recommended in the conclusion is compatible with either reading: it asks only that a circuit-level claim state the level of description it rests on.

\paragraph{Broader impact.}
The setting consists of controlled transformers trained on synthetic Lean tactic-prediction tasks, not deployed systems. The practical benefit is methodological: studies that name which graph is reported, how it is extracted, and which comparison supports a mechanistic conclusion are better positioned to produce reproducible and comparable circuit analyses. The corresponding indirect risk is over-interpretation: findings from a small synthetic benchmark, or weakly validated mechanistic claims based on a single extraction recipe, could mislead downstream users if treated as guidance for deployed large models. The reporting practice recommended in the conclusion is intended to mitigate this risk, and extending the controlled comparison to deployed-model interpretability would require validation across larger models, longer contexts, and more heterogeneous tasks.

\FloatBarrier

\section{Reproducibility Package and Compute Resources}
\label{app:reprocompute}

Our code is publicly available at \url{https://github.com/Stepuuu/circuit-extraction-stability}. That release supports the training and circuit-extraction experiments and contains the synthetic data-generation and tokenizer scripts, supervised and GRPO training entry points, task-level evaluation scripts, circuit-extraction scripts for the core and factorized query/key graph objects, fixed-loss pruning sweep code, a minimal runtime module, a requirements file, and representative commands. The reported experiments do not redistribute third-party datasets or pretrained models as experimental assets.

The released training and analysis scripts run supervised, GRPO, and circuit-analysis workloads as independent GPU-accelerated jobs. For the dominant supervised recipe in Table~\ref{tab:training_recipes}, a 5B-token run of the shared 8-layer, 415.7M-parameter transformer corresponds to approximately $1.2\times 10^{19}$ training FLOPs under the standard dense-transformer $6ND$ estimate \citep{Kaplan2020Scaling,Hoffmann2022Chinchilla}. The paper and supplemental package specify the model size, token budgets, batch sizes, gradient-accumulation settings, sequence length, evaluation cadence, and RL sampling settings that determine the training and analysis workload scale.

\FloatBarrier
\section{Additional protocol and causal checks}
\label{app:rebuttal_audits}

These checks distinguish five questions: sensitivity to ranking conventions,
reproducibility within a checkpoint, comparison across training stages and depths,
selectivity of bridge-node interventions, and agreement with independent causal
measurements.

\subsection{Ranking conventions and random baselines}
\label{app:rebuttal-ranking}
Routing scores count the edges assigned to each attention head. Many heads receive the same score, and deterministic selection resolves ties using head enumeration order. At the dense--$75\%$-sparse RL comparison, head Jaccard@10 is $0.666/0.554$ for the success/near-miss splits. Random selection within tied score groups gives approximately $0.028/0.027$ while keeping the extracted graphs and scores fixed.

Cross-checkpoint comparisons match entries by task, extracted object, and generation seed. The intervention and SFT/depth comparisons use the same frozen inputs at both endpoints.

For structural overlap, Table~\ref{tab:rebuttal-null} compares uniform and score-weighted random baselines. Both sample without replacement from the split-level union of observed top-$k$ structural edge bundles, preserving the observed set sizes. Uniform sampling assigns equal probability to these bundles. Score-weighted sampling uses each checkpoint's reported bundle scores; a bundle absent from that checkpoint's candidate set receives its minimum reported weight, bounded below by $10^{-12}$. For each of 2,000 Monte Carlo draws, we average Jaccard over the 16 matched task--object entries in the split. The table reports the baseline means and one-sided tail probabilities. The observed structural overlap is close to the score-weighted baseline in both splits.

\begin{table}[htbp]
\centering\small
\caption{Structural overlap and random baselines for the dense and $75\%$ weight-sparse RL checkpoints. Each observed value is mean Jaccard@10 over 16 matched task--object entries per split. Size null samples uniformly while preserving the observed set sizes; Score-weighted null samples without replacement using the scores at each checkpoint. Both use the split-level union of observed top-$k$ structural edge bundles and 2,000 Monte Carlo draws. The $p$ columns report $(1+\#\{J_{\rm null}\geq J_{\rm obs}\})/2001$.}
\label{tab:rebuttal-null}
\begin{tabular}{lrrrrr}
\toprule
Split & Observed $J@10$ & Size null & $p$ & Score-weighted null & $p$ \\
\midrule
Success & 0.1631 & 0.1261 & 0.040 & 0.1548 & 0.319 \\
Near-miss & 0.1423 & 0.1166 & 0.100 & 0.1367 & 0.387 \\
\bottomrule
\end{tabular}
\end{table}


Table~\ref{tab:additional-overlap-nulls} separates overlap induced by head aggregation from structural comparisons at other sparsity levels. For the head-aggregation baseline, each candidate head contributes its edge count of pseudo-edges, with a minimum of one. We sample up to ten pseudo-edges without replacement at each endpoint, group them by attention head, and compute Jaccard. This baseline measures overlap induced by head multiplicity and aggregation.

The structural rows use the score-weighted sampling procedure above. Each of 2,000 draws averages Jaccard over the matched task--object entries for that comparison. The head comparison has 16 entries per split; each additional structural comparison has 18 success entries at training seed 123.

\begin{table}[htbp]
\centering\small
\caption{Additional overlap baselines for the comparisons in Section~\ref{sec:routing_anchor}. The routing baseline samples pseudo-edges according to head edge counts and groups the sampled edges by head before computing Jaccard; routing observations use deterministic selection. The structural rows compare other sparsity levels using score-weighted random sampling. Here $n$ counts matched task--object entries, and Null mean averages 2,000 Monte Carlo draws.}
\label{tab:additional-overlap-nulls}
\begin{tabular}{lrrr}
\toprule
Comparison & $n$ & Observed $J$ & Null mean \\
\midrule
\multicolumn{4}{l}{\textit{Routing heads: coarsening-only null}} \\
Dense--75\%, success & 16 & 0.666 & 0.250 \\
Dense--75\%, near-miss & 16 & 0.554 & 0.225 \\
\addlinespace
\multicolumn{4}{l}{\textit{Structural edge bundles: per-side score-weighted null}} \\
Dense--25\%, success & 18 & 0.287 & 0.215 \\
Dense--50\%, success & 18 & 0.326 & 0.186 \\
Dense--90\%, success & 18 & 0.167 & 0.109 \\
\bottomrule
\end{tabular}
\end{table}

\subsection{Repeated circuit extraction}
\label{app:rebuttal-repeat}
We repeat extraction with two different extraction seeds at each seed-42 RL checkpoint, keeping the extraction implementation and prompt-seed selections fixed. Each setting contains 18 entries: nine tasks for each of the extended support and fixed-loss pruning graphs. We compare their structural edge-bundle rankings and top-ten sets.

The size-matched baseline samples up to ten structural edge bundles from each endpoint's own candidate set, using 2,000 draws per entry. Rank-biased overlap (RBO, persistence $0.9$) compares the rankings and is normalized by their finite shared depth. Its baseline uses 200 independent permutations of the candidate rankings. Table~\ref{tab:rebuttal-repeat} reports both measures, their baselines, and the number of entries with unadjusted $p<0.05$.

\begin{table}[htbp]
\centering\small
\caption{Structural overlap across repeated extractions of the same RL checkpoint. Each row averages 18 task--object entries. Null is the mean size-matched random baseline, and Excess is observed Jaccard@10 minus that mean. The $p<.05$ column counts entries with unadjusted tail probabilities below 0.05. RBO denotes rank-biased overlap; RBO null is its mean under random permutations of the candidate rankings.}
\label{tab:rebuttal-repeat}
\begin{tabular}{llrrrrrr}
\toprule
Checkpoint & Split & $J@10$ & Null & Excess & $p<.05$ & RBO & RBO null \\
\midrule
Dense & Success & 0.7853 & 0.6140 & +0.1713 & 6/18 & 0.8752 & 0.3903 \\
Dense & Near-miss & 0.7821 & 0.6022 & +0.1798 & 8/18 & 0.8787 & 0.3956 \\
75\%-sparse & Success & 0.8024 & 0.7736 & +0.0288 & 0/18 & 0.8823 & 0.4102 \\
75\%-sparse & Near-miss & 0.7897 & 0.7542 & +0.0355 & 1/18 & 0.8563 & 0.4012 \\
\bottomrule
\end{tabular}
\end{table}

Repeated-extraction Jaccard is $0.78$--$0.80$, and RBO is $0.86$--$0.88$. The sparse-checkpoint Jaccard baseline is also high ($0.75$--$0.77$), whereas ranking agreement remains higher than its permutation baseline. These experiments measure repeatability within a checkpoint. The cross-checkpoint comparisons use training seed 123 and differ in extraction implementation and task--object coverage.

\subsection{SFT and reduced-depth comparisons}
\label{app:rebuttal-stage-depth}
We compare dense and $75\%$-sparse SFT checkpoints at eight and four layers, using training seed 42 and the same synthetic task family. The checkpoints retain the RMSNorm normalization used during SFT training. We extract all three circuit objects and evaluate behavior with the SFT models.

The extraction recipe uses 150 epochs, 100 optimization examples per task, batch size 32, and 200 reference candidates per task. We retain the object-specific objectives and thresholds, set an optimizer seed for each task and object, and freeze the generated inputs for paired comparisons. Edge scoring enumerates connections while retaining the original scores and stable ordering of the top 200 edges; node counts and nonzero-edge counts cover the complete selected graph.

Input groups A and B use the generation-seed lists associated with the success and near-miss comparisons, respectively. Their frozen prompts are shared between the dense and sparse endpoints. Each group has 18 entries: nine tasks for each of the extended support and fixed-loss pruning graphs. Table~\ref{tab:rebuttal-stage-depth} compares structural overlap with the per-entry size-matched baseline defined in Section~\ref{app:rebuttal-repeat}. Routing scores retain ties, so the head-selection sensitivity described in Section~\ref{app:rebuttal-ranking} also applies.

\begin{table}[htbp]
\centering\small
\caption{Structural Jaccard@10 between dense and $75\%$ weight-sparse SFT checkpoints at two model depths. A and B denote the two fixed input groups defined in Section~\ref{app:rebuttal-stage-depth}. Each row averages 18 entries: nine tasks for each of the extended support and fixed-loss pruning graphs. Size null is the mean size-matched random baseline; Excess is observed overlap minus that mean. The final column counts entries with unadjusted tail probabilities below 0.05.}
\label{tab:rebuttal-stage-depth}
\begin{tabular}{llrrrr}
\toprule
Setting & Prompt set & $J@10$ & Size null & Excess & $p<.05$ \\
\midrule
8-layer SFT & A & 0.1371 & 0.1391 & -0.0020 & 0/18 \\
8-layer SFT & B & 0.1510 & 0.1655 & -0.0145 & 0/18 \\
4-layer SFT & A & 0.4075 & 0.3947 & +0.0128 & 0/18 \\
4-layer SFT & B & 0.4032 & 0.3842 & +0.0190 & 0/18 \\
\bottomrule
\end{tabular}
\end{table}

Structural Jaccard@10 is close to its size-matched baseline at both depths. None of the individual task--object comparisons has an unadjusted $p<0.05$. The larger raw overlap at four layers is accompanied by a larger random-baseline overlap.

We also compare coupled and factorized Q/K representations for the extended support and fixed-loss pruning graphs at 12 SFT checkpoints: two dense, three $25\%$-sparse, three $50\%$-sparse, two $75\%$-sparse, and two $90\%$-sparse models. We use the extraction recipes in Table~\ref{tab:extraction_recipes}; the core circuit has no Q/K split. Each checkpoint contributes five compositional tasks. We average node fractions over tasks, take the absolute difference between representations, and then average over checkpoints. The mean difference is $0.00709$ for the extended support graph (maximum $0.01520$) and $0.01396$ for the fixed-loss pruning graph (maximum $0.08045$). Pruning uses a grid-plus-local search over the fitted score ranking at $\tau=0.12$; all 120 pruning cells meet the budget.

The extended support graph preserves all ten pairwise sparsity-condition orderings; the fixed-loss pruning graph preserves nine. For pruning, the $25\%$ and $90\%$ means are $0.20664$ and $0.18525$ with coupled Q/K, versus $0.17330$ and $0.17361$ with factorized Q/K. The largest checkpoint gap occurs at $25\%$ sparsity, seed 456. The 12 checkpoints form an incomplete sparsity--seed grid.

Reference means average batch means. Support fitting uses batches of 16 for both representations; pruning fitting uses 16 for coupled Q/K and 32 for factorized Q/K. Common-budget reevaluation uses 32 for both. Paired runs share optimization and reference inputs; final support and pruning evaluations also match reference means within their respective protocols.

\subsection{Bridge-node interventions}
\label{app:rebuttal-bridge}
For each compositional task, bridge nodes are the attention-output channels and MLP neurons in its core circuit that lie outside the union of the corresponding atomic-task circuits. We treat each attention-output channel and each MLP neuron as a distinct node, and replace bridge-node activations with reference means while keeping other nodes intact.

For these comparisons, \textsc{AND-OR}, \textsc{OR-AND}, and \textsc{AND-OR-AND} use the \textsc{AND} and \textsc{OR-L} atomic circuits; \textsc{Nested-AND} uses \textsc{AND} and \textsc{ID}; and \textsc{Nested-OR} uses \textsc{OR-L} and \textsc{ID}. Here \textsc{ID} is the identity task. Each atomic circuit enters the set union once.

We evaluate dense and $75\%$-sparse RL and SFT checkpoints, all with training seed 42. The RL pair uses the narrow baseline training distribution, and the SFT pair uses the moderately expanded distribution. We use the RL core selections from the graph analysis and extract the SFT core circuits from the RMSNorm models with the supplementary recipe in Section~\ref{app:rebuttal-stage-depth}. Every checkpoint is evaluated on the same 200 frozen examples per task. We measure teacher-forced tactic-output exact match (EM), which differs from autoregressive generation accuracy.

At the $75\%$-sparse RL checkpoint, ablating the full bridge set reduces compositional EM on all five tasks, while EM on the corresponding atomic tasks remains between $0.995$ and $1.000$. Effects differ across the other checkpoints, including tasks with little or no compositional decrease. Table~\ref{tab:rebuttal-bridge} reports all 20 checkpoint--task combinations.

For comparisons with non-bridge nodes, we sample without replacement. Bridge and non-bridge subsets are matched in node count within node family, layer, and selection-score bin, as described in Appendix~\ref{app:bridge-confirmation}. These matched subsets are evaluated separately from the full bridge sets. Table~\ref{tab:rebuttal-bridge} reports full-set effects and the matched subset sizes.

\begin{table}[htbp]
\caption{Full bridge-node ablation across four checkpoints and five compositional tasks. Values are teacher-forced tactic-output exact match on the same 200 examples per task. C denotes the compositional task; A gives the range across its two corresponding atomic tasks after full bridge ablation. $N$ is the full bridge-set size. $n$ is the number of nodes ablated in each of the bridge and non-bridge subsets matched by node family, layer, and selection-score bin (Appendix~\ref{app:bridge-confirmation}); matching uses their common support and can yield smaller subsets. Checkpoint configurations are specified in Section~\ref{app:rebuttal-bridge}.}
\label{tab:rebuttal-bridge}
\centering
\scriptsize
\setlength{\tabcolsep}{3pt}
\renewcommand{\arraystretch}{1.08}
\begin{tabularx}{\linewidth}{l l c c >{\centering\arraybackslash}X r r}
\toprule
Checkpoint & Task & \shortstack{Baseline\\C} & \shortstack{Full bridge\\C} & \shortstack{Full bridge\\A} & $N$ & $n$ \\
\midrule
RL dense & \texttt{T\_and\_or} & 0.535 & 0.535 & 1.000 & 53,950 & 12,023 \\
RL dense & \texttt{T\_or\_and} & 0.990 & 0.770 & 1.000 & 3,463 & 3,290 \\
RL dense & \texttt{T\_and\_or\_and} & 0.535 & 0.045 & 0.995 & 51,407 & 11,585 \\
RL dense & \texttt{T\_nested\_and} & 0.995 & 0.990 & 1.000 & 2,694 & 2,517 \\
RL dense & \texttt{T\_nested\_or} & 0.300 & 0.300 & 0.975--1.000 & 54,391 & 10,088 \\
\midrule
RL $75\%$ sparse & \texttt{T\_and\_or} & 1.000 & 0.930 & 1.000 & 3,726 & 3,102 \\
RL $75\%$ sparse & \texttt{T\_or\_and} & 0.995 & 0.940 & 0.995--1.000 & 3,080 & 2,914 \\
RL $75\%$ sparse & \texttt{T\_and\_or\_and} & 0.990 & 0.460 & 0.995--1.000 & 20,415 & 5,959 \\
RL $75\%$ sparse & \texttt{T\_nested\_and} & 0.995 & 0.405 & 1.000 & 6,675 & 4,115 \\
RL $75\%$ sparse & \texttt{T\_nested\_or} & 0.975 & 0.295 & 1.000 & 20,509 & 5,010 \\
\midrule
SFT dense & \texttt{T\_and\_or} & 0.535 & 0.535 & 1.000 & 49,228 & 17,322 \\
SFT dense & \texttt{T\_or\_and} & 1.000 & 0.620 & 1.000 & 3,850 & 3,734 \\
SFT dense & \texttt{T\_and\_or\_and} & 0.535 & 0.360 & 1.000 & 46,898 & 16,983 \\
SFT dense & \texttt{T\_nested\_and} & 0.995 & 0.000 & 1.000 & 12,286 & 6,081 \\
SFT dense & \texttt{T\_nested\_or} & 0.305 & 0.300 & 1.000 & 50,724 & 16,861 \\
\midrule
SFT $75\%$ sparse & \texttt{T\_and\_or} & 0.855 & 0.855 & 1.000 & 27,581 & 25,403 \\
SFT $75\%$ sparse & \texttt{T\_or\_and} & 0.995 & 0.985 & 0.995--1.000 & 6,752 & 6,725 \\
SFT $75\%$ sparse & \texttt{T\_and\_or\_and} & 0.850 & 0.845 & 1.000 & 28,560 & 25,839 \\
SFT $75\%$ sparse & \texttt{T\_nested\_and} & 0.995 & 0.000 & 1.000 & 18,013 & 6,712 \\
SFT $75\%$ sparse & \texttt{T\_nested\_or} & 0.700 & 0.700 & 1.000 & 30,362 & 26,630 \\
\bottomrule
\end{tabularx}
\end{table}

\FloatBarrier
\subsection{Single-head intervention checks}
\label{app:rebuttal-node-probes}
At the seed-42 dense and $75\%$-sparse RL checkpoints, we intervene on each of the 1,024 attention heads separately for each of the nine tasks. Mean ablation replaces a head's output with a reference mean; activation patching replaces it with activations from a counterfactual prompt. Counterfactual prompts have the same input and total token lengths as their clean partners, with different correct answers preferred when constructing pairs. We measure changes in cross-entropy loss on the clean target tokens.

For each task, we compare the top-ten heads selected by routing scores for the cross-checkpoint comparison with the 100 heads having the largest measured loss effects. This gives 90 task--head selections per checkpoint; a head can occur in more than one task. Table~\ref{tab:rebuttal-node-probes} reports their overlap, alongside a uniform-head reference.

\begin{table}[htbp]
\centering\small
\caption{Agreement between routing-selected heads and intervention-based rankings. Entries count how many of the ten routing-selected heads per task also appear among the 100 heads with the largest probe loss effects. Across nine tasks there are 90 task--head selections per checkpoint; the same head can count once in each task. Uniform head selection gives an expected count of $90(100/1024)=8.79$.}
\label{tab:rebuttal-node-probes}
\begin{tabular}{lrr}
\toprule
Probe & Dense & 75\%-sparse \\
\midrule
Mean ablation & 31/90 & 46/90 \\
Counterfactual patching & 32/90 & 39/90 \\
\bottomrule
\end{tabular}
\end{table}

The selected heads follow the deterministic routing rule and are concentrated in particular layers; the table uses uniform sampling over all heads as its reference. Ranking heads directly by their measured intervention effects gives a dense--sparse top-ten Jaccard of $0.00585$ for both mean ablation and activation patching.

We also compare head-gate gradient and integrated-gradient estimates with these measured effects to assess the estimators. Appendix~\ref{app:weight_edge_audit} defines the separate path-integrated weight attribution method and its weight-level intervention check.

\FloatBarrier
\section{Path-integrated weight attribution}
\label{app:weight_edge_audit}

\paragraph{Purpose.}
The three graph objects in the main study use related masking and pruning procedures. To compare circuits under a different score, we adapt the path-integration approach of EAP-IG~\citep{Hanna2024Faithfulness} to individual weight connections. This adaptation scores entries of weight matrices along a path that jointly changes several internal activations.

\paragraph{Inputs, path, and score.}
We use the seed-42 dense and $75\%$-sparse RL checkpoints, all nine tasks, and 64 clean--counterfactual pairs per task and checkpoint. Each counterfactual has the same input-token and full-sequence lengths as its clean partner. The two prompts in each pair have different expected tactics. The model runs in BF16; score accumulation uses float32, with 20 midpoint integration steps and batches of eight. The structural-overlap comparison uses training seed 123; this analysis uses seed 42.

The scored matrices are the query/key/value and output projections of attention, and the input and output projections of each MLP, in all eight layers. For a weight $w_{ji}$ connecting source channel $i$ to linear-output channel $j$, let $\Delta s_i$ be the clean-minus-counterfactual source activation captured from ordinary forward passes. At integration point $\alpha_m=(m-\tfrac12)/20$, embeddings interpolate between their captured endpoints. At each subsequent scored source, the source activation is $s^{\mathrm{cf}}+\alpha_m(s^{\mathrm{current}}(\alpha_m)-s^{\mathrm{cf}})$, retaining the dependence on the upstream computation. Its signed score is
\begin{equation}
 A_{ji}=\frac{w_{ji}}{20N}\sum_{m=1}^{20}\sum_{b,t}
 \frac{\partial\mathcal{L}_{\mathrm{sum}}(\alpha_m)}{\partial z_{btj}(\alpha_m)}
 \Delta s_{bti},
\label{eq:weight_edge_audit_score}
\end{equation}
where $N$ is the total number of supervised target tokens and $\mathcal{L}_{\mathrm{sum}}$ is their summed cross-entropy loss against the clean targets. This score uses the fixed clean-minus-counterfactual displacement $\Delta s_i$, rather than the derivative of the jointly patched source along the path; an integrated-gradients completeness identity is therefore not assumed. We rank weight edges by $|A_{ji}|$. Head scores sum the absolute scores of the associated query/key/value and attention-output weights; MLP weights contribute only to the weight-edge ranking.

The clean input's direct bigram-logit term is held fixed along the path. The $\alpha=0$ endpoint combines counterfactual internal activations with this clean direct term; $\alpha=1$ recovers the clean forward pass.

\paragraph{Comparison and null distributions.}
We compare the top 10 heads and top 10 weight edges across checkpoints for each task. No head or weight-edge top-ten cutoff contains a tie in the 18 task--checkpoint conditions. The head null draws two independent sets of 10 from all $8\times128=1024$ heads. The edge null first samples a matrix according to each checkpoint's total absolute attribution mass in that matrix, then samples a coordinate uniformly within its full dimensions. Duplicate coordinates are excluded within each sampled set. This baseline preserves matrix-level attribution mass and samples from all matrix coordinates, including zero-weight positions. We use 5000 Monte Carlo draws with random seed 20260803; each aggregate draw averages the nine task overlaps. Reported tail probabilities use $(1+\#\{J_{\mathrm{null}}\geq J_{\mathrm{obs}}\})/5001$.

Mean head Jaccard@10 is $0.005848$, against a uniform-head null mean of $0.005343$ ($p=0.6001$). Individual weight-edge Jaccard@10 is zero on all nine tasks, as is the sampled edge-null mean ($p=1$). The selected top-ten weight sets are disjoint; the sampled baseline also has zero overlap in this large coordinate universe. 

\paragraph{Weight-ablation check.}
We zero the 10,000 highest-ranked weights and measure the change in clean-target cross entropy per target token on the same 64 examples used for attribution. For each condition, three control sets contain the same number of nonzero weights from each selected matrix, sampled uniformly within that matrix's nonzero support. Each intervention starts from the unmodified checkpoint. Table~\ref{tab:weight_edge_interventions} compares the loss change with the mean of these three controls. Top-edge ablation increases loss more than the control mean in 16 of 18 conditions. In both \texttt{T\_and\_or\_and} conditions, it decreases loss and falls below the control mean. Ranking by absolute attribution can select connections with opposing signed effects.

\begin{table}[htbp]
\centering
\small
\setlength{\tabcolsep}{4pt}
\begin{tabular}{llrrrr}
\toprule
Checkpoint & Task & Pairs & Top-edge $\Delta\ell$ & Random $\Delta\ell$ & Top $>$ random \\
\midrule
Dense & \texttt{S\_and} & 64 & $1.724\!\times\!10^{-4}$ & $-1.106\!\times\!10^{-7}$ & Yes \\
Dense & \texttt{S\_or\_l} & 64 & $2.504\!\times\!10^{-4}$ & $-4.565\!\times\!10^{-6}$ & Yes \\
Dense & \texttt{S\_or\_r} & 64 & 0.58139 & $3.935\!\times\!10^{-7}$ & Yes \\
Dense & \texttt{S\_id} & 64 & $1.424\!\times\!10^{-4}$ & $5.369\!\times\!10^{-7}$ & Yes \\
Dense & \texttt{T\_and\_or} & 64 & 1.20985 & $5.336\!\times\!10^{-4}$ & Yes \\
Dense & \texttt{T\_or\_and} & 64 & 0.00653 & $7.695\!\times\!10^{-6}$ & Yes \\
Dense & \texttt{T\_and\_or\_and} & 64 & -0.03072 & $-4.598\!\times\!10^{-4}$ & No \\
Dense & \texttt{T\_nested\_and} & 64 & 0.00188 & $1.357\!\times\!10^{-6}$ & Yes \\
Dense & \texttt{T\_nested\_or} & 64 & 0.44583 & $-5.934\!\times\!10^{-4}$ & Yes \\
$75\%$ sparse & \texttt{S\_and} & 64 & 1.22397 & $5.344\!\times\!10^{-7}$ & Yes \\
$75\%$ sparse & \texttt{S\_or\_l} & 64 & 0.04427 & $1.330\!\times\!10^{-6}$ & Yes \\
$75\%$ sparse & \texttt{S\_or\_r} & 64 & 0.14806 & $-1.722\!\times\!10^{-7}$ & Yes \\
$75\%$ sparse & \texttt{S\_id} & 64 & 0.11376 & $-4.602\!\times\!10^{-7}$ & Yes \\
$75\%$ sparse & \texttt{T\_and\_or} & 64 & 0.30923 & $-8.034\!\times\!10^{-7}$ & Yes \\
$75\%$ sparse & \texttt{T\_or\_and} & 64 & 0.10780 & $-5.966\!\times\!10^{-7}$ & Yes \\
$75\%$ sparse & \texttt{T\_and\_or\_and} & 64 & $-8.283\!\times\!10^{-4}$ & $-5.384\!\times\!10^{-5}$ & No \\
$75\%$ sparse & \texttt{T\_nested\_and} & 64 & 0.38530 & $-8.219\!\times\!10^{-6}$ & Yes \\
$75\%$ sparse & \texttt{T\_nested\_or} & 64 & 0.07218 & $-9.814\!\times\!10^{-5}$ & Yes \\
\bottomrule
\end{tabular}
\caption{Weight-attribution intervention results for all 18 task--checkpoint conditions. $\Delta\ell$ is the change in clean-target cross entropy per supervised token after zeroing the 10,000 highest-ranked weights. Random is the mean over three controls matched by matrix and restricted to nonzero weights. Negative values indicate a loss decrease after ablation. Top $>$ random compares the loss change with the control mean.}
\label{tab:weight_edge_interventions}
\end{table}

\section{Matched controls for bridge ablation}
\label{app:bridge-confirmation}
We compare the full bridge-set intervention with equal-sized bridge and non-bridge subsets to examine how task selectivity depends on bridge membership. We use three intervention groups: the full bridge set, a matched bridge subset, and a matched non-bridge subset.

\paragraph{Matched subsets.}
We stratify candidate nodes by node family and layer. Within each stratum, we sample equal numbers of bridge and non-bridge nodes, using the size of the smaller pool. A second matching rule additionally stratifies by the learned selection score in fixed-width bins of $0.1$. All sampling is without replacement. Matching can yield subsets substantially smaller than the full bridge set when non-bridge candidates are scarce.

For each task and checkpoint, we use three paired mask draws fixed before evaluation.

\paragraph{Evaluation and task selectivity.}
Each intervention is evaluated on 200 examples from the compositional task and 200 from each of its two atomic tasks. Examples are identical across interventions within a task. We measure teacher-forced tactic-output exact match (EM): every scored output token must be correct given its ground-truth prefix. Reference activations are computed by averaging batch means over 200 examples from each of the nine tasks.

\begin{figure}[htbp]
\centering
\includegraphics[width=\linewidth]{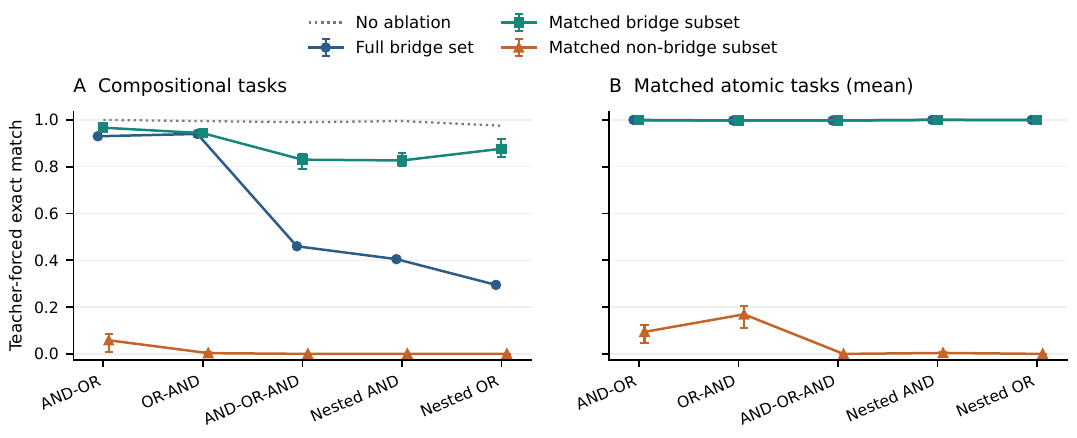}
\caption{Full bridge-set ablation and matched-subset ablation at the $75\%$-sparse RL checkpoint described in Section~\ref{app:rebuttal-bridge}. Left: compositional-task exact match (EM). Right: mean EM over the two corresponding atomic tasks. Matched-subset points average three paired mask draws; error bars span their minimum and maximum. Matched bridge and non-bridge subsets have equal node counts within each node-family, layer, and selection-score bin. Full bridge-set ablation removes the entire bridge set.}
\label{fig:bridge-matched-confirmation}
\end{figure}

Let $\Delta_C$ be the compositional EM drop from the unablated model, and $\Delta_{A_1},\Delta_{A_2}$ the corresponding atomic drops. Task selectivity and relative task selectivity are
\[
 S=\Delta_C-\tfrac12(\Delta_{A_1}+\Delta_{A_2}),
 \qquad R=S_{\mathrm{bridge}}-S_{\mathrm{nonbridge}}.
\]
A positive $S$ indicates greater impairment on the compositional task than on its atomic constituents. A positive $R$ indicates greater task selectivity for the matched bridge subset than for the matched non-bridge subset. We report raw EM alongside these contrasts to show the absolute effects of each intervention (Figure~\ref{fig:bridge-matched-confirmation}; Table~\ref{tab:bridge-confirmation-main}). EM is reported as a proportion; relative selectivity is reported in percentage points.

We report all task--checkpoint combinations. At the $75\%$-sparse RL checkpoint, full bridge removal impairs all five compositional tasks while largely preserving their atomic constituents. Relative selectivity of the matched subsets varies across tasks. In some tasks, non-bridge removal impairs both compositional and atomic performance, yielding a positive relative selectivity even when the absolute bridge effect is small. These fixed-checkpoint interventions complement the association between bridge fraction and compositional gains across training conditions.

\begin{table}[htbp]
\caption{Exact-match accuracy under full bridge-set and matched-subset ablation at the $75\%$-sparse RL checkpoint. Values are teacher-forced tactic-output EM on 200 examples per task. C denotes the compositional task and A its two atomic constituents; atomic baselines are $1.000$. Full-bridge columns report compositional EM and the atomic EM range after ablating the entire bridge set. Matched bridge and non-bridge subsets have equal counts within each node-family, layer, and selection-score bin; $n$ is the number ablated in each subset. Matched columns give compositional and atomic EM ranges over three paired mask draws, with atomic ranges also spanning the two atomic tasks.}
\label{tab:bridge-confirmation-main}
\centering
\scriptsize
\setlength{\tabcolsep}{3pt}
\renewcommand{\arraystretch}{1.12}
\begin{tabularx}{\linewidth}{l c c c r >{\centering\arraybackslash}X >{\centering\arraybackslash}X}
\toprule
Task & \shortstack{Baseline\\C} & \shortstack{Full bridge\\C} & \shortstack{Full bridge\\A} & $n$ & \shortstack{Matched bridge\\C; A} & \shortstack{Matched non-bridge\\C; A} \\
\midrule
\texttt{T\_and\_or} & 1.000 & 0.930 & 1.000 & 3,102 & 0.965--0.970; 0.995--1.000 & 0.010--0.085; 0.000--0.250 \\
\texttt{T\_or\_and} & 0.995 & 0.940 & 0.995--1.000 & 2,914 & 0.940--0.945; 0.995--1.000 & 0.000--0.010; 0.000--0.410 \\
\texttt{T\_and\_or\_and} & 0.990 & 0.460 & 0.995--1.000 & 5,959 & 0.790--0.855; 0.995--1.000 & 0.000; 0.000 \\
\texttt{T\_nested\_and} & 0.995 & 0.405 & 1.000 & 4,115 & 0.805--0.860; 1.000 & 0.000; 0.000--0.015 \\
\texttt{T\_nested\_or} & 0.975 & 0.295 & 1.000 & 5,010 & 0.840--0.920; 0.995--1.000 & 0.000; 0.000 \\
\bottomrule
\end{tabularx}
\end{table}

\begin{table}[htbp]
\caption{Relative task selectivity $R$ of matched ablations across four checkpoints and five compositional tasks. Checkpoint configurations are given in Section~\ref{app:rebuttal-bridge}; $R$ is defined in Appendix~\ref{app:bridge-confirmation}. Entries report the mean over three paired mask draws in percentage points, with pointwise paired-bootstrap 95\% intervals (2,000 replicates). Within each task, the same resampled example indices are used across intervention conditions; tasks are resampled independently. Checkpoints, circuit selections, and masks remain fixed during resampling. The two columns match node family and layer, or additionally match selection scores in bins of width $0.1$.}
\label{tab:bridge-selectivity}
\centering
\scriptsize
\setlength{\tabcolsep}{4pt}
\renewcommand{\arraystretch}{1.08}
\begin{tabularx}{\linewidth}{l l >{\centering\arraybackslash}X >{\centering\arraybackslash}X}
\toprule
Checkpoint & Task & Family/layer & Family/layer/score bin \\
\midrule
RL dense & \texttt{T\_and\_or} & +46.50 [+39.00, +53.50] & +46.50 [+39.00, +53.50] \\
RL dense & \texttt{T\_or\_and} & +16.58 [+11.17, +22.42] & +15.83 [+10.25, +21.50] \\
RL dense & \texttt{T\_and\_or\_and} & +48.33 [+41.17, +55.33] & +48.17 [+41.00, +55.17] \\
RL dense & \texttt{T\_nested\_and} & +0.92 [-0.17, +2.50] & +0.75 [-0.33, +2.33] \\
RL dense & \texttt{T\_nested\_or} & +70.00 [+63.50, +76.00] & +70.00 [+63.50, +76.00] \\
\midrule
RL $75\%$ sparse & \texttt{T\_and\_or} & +1.67 [-2.33, +6.00] & -0.42 [-4.17, +3.50] \\
RL $75\%$ sparse & \texttt{T\_or\_and} & -7.58 [-11.42, -3.50] & -11.17 [-15.33, -6.83] \\
RL $75\%$ sparse & \texttt{T\_and\_or\_and} & +15.42 [+10.75, +19.83] & +16.75 [+11.92, +21.25] \\
RL $75\%$ sparse & \texttt{T\_nested\_and} & +15.00 [+10.58, +19.67] & +16.92 [+12.08, +21.92] \\
RL $75\%$ sparse & \texttt{T\_nested\_or} & +8.92 [+5.42, +12.58] & +12.17 [+8.33, +16.33] \\
\midrule
SFT dense & \texttt{T\_and\_or} & +46.50 [+39.50, +53.00] & +46.50 [+39.50, +53.00] \\
SFT dense & \texttt{T\_or\_and} & +6.67 [-0.75, +14.50] & +5.75 [-1.67, +13.84] \\
SFT dense & \texttt{T\_and\_or\_and} & +46.75 [+40.00, +53.42] & +46.92 [+40.25, +53.59] \\
SFT dense & \texttt{T\_nested\_and} & +80.83 [+75.83, +85.34] & +70.17 [+64.33, +75.67] \\
SFT dense & \texttt{T\_nested\_or} & +70.00 [+63.50, +76.00] & +70.00 [+63.50, +76.00] \\
\midrule
SFT $75\%$ sparse & \texttt{T\_and\_or} & +14.50 [+10.00, +19.00] & +14.50 [+10.00, +19.00] \\
SFT $75\%$ sparse & \texttt{T\_or\_and} & -26.58 [-32.33, -20.91] & -22.50 [-28.75, -16.42] \\
SFT $75\%$ sparse & \texttt{T\_and\_or\_and} & +15.33 [+10.50, +20.34] & +15.33 [+10.50, +20.34] \\
SFT $75\%$ sparse & \texttt{T\_nested\_and} & +30.67 [+25.00, +36.33] & +58.67 [+53.00, +64.17] \\
SFT $75\%$ sparse & \texttt{T\_nested\_or} & +30.00 [+24.00, +36.00] & +30.00 [+24.00, +36.00] \\
\bottomrule
\end{tabularx}
\end{table}

\FloatBarrier

\end{document}